\PassOptionsToPackage{dvipsnames}{xcolor}
\documentclass{article}

\usepackage{microtype}
\usepackage{graphicx}
\usepackage{subfigure}
\usepackage{booktabs} %

\usepackage{hyperref}

\usepackage[accepted]{icml2024}

\usepackage{amsmath}
\usepackage{amssymb}
\usepackage{mathtools}
\usepackage{amsthm}

\usepackage[capitalize,noabbrev]{cleveref}

\theoremstyle{plain}
\newtheorem{theorem}{Theorem}[section]

\newtheorem{lemma}[theorem]{Lemma}

\theoremstyle{definition}

\theoremstyle{remark}

\usepackage[textsize=tiny]{todonotes}
\usepackage{fancyhdr}
\usepackage{xcolor}
\usepackage{algorithm}
\usepackage{algorithmic}
\usepackage{natbib}

\usepackage{scalerel}
\usepackage{float}

\usepackage{subcaption}
\usepackage{caption}
\usepackage[toc,page]{appendix}
\usepackage{url}            %
\usepackage{amsfonts}       %
\usepackage{nicefrac}       %
\usepackage{cancel}
\usepackage{latexsym}
\usepackage[most]{tcolorbox}
\usepackage{placeins}
\usepackage{bold-extra}
\usepackage[T1]{fontenc}
\usepackage{standalone}
\usepackage{pgf}
\usepackage[customcolors,shade]{hf-tikz}
\usepackage{nccmath}
\usepackage{multirow}
\usepackage{stfloats}
\usepackage{xargs}
\usepackage{float}
\usepackage{mdframed}
\usepackage{comment}
\usepackage{enumerate}
\usepackage{authblk}

\usepackage[normalem]{ulem}

\newcommand{\sgparam}{\tau}

\newcommand{\gparam}{\boldsymbol{\tau}}

\newcommand{\MExp}{\mathrm{Expm}}

\newcommand{\stepsize}{\beta}

\newcommand{\fim}{F}

\newcommand\cut[1]{}

\newcommand{\elbofinal}{\mathcal{L}}

\newcommand{\squishlist}{
   \begin{list}{$\bullet$}
    { \setlength{\itemsep}{0pt}      \setlength{\parsep}{3pt}
      \setlength{\topsep}{3pt}       \setlength{\partopsep}{0pt}
      \setlength{\leftmargin}{1.5em} \setlength{\labelwidth}{1em}
      \setlength{\labelsep}{0.5em} } }

\newcommand{\squishlisttwo}{
   \begin{list}{$\bullet$}
    { \setlength{\itemsep}{0pt}    \setlength{\parsep}{0pt}
      \setlength{\topsep}{0pt}     \setlength{\partopsep}{0pt}
      \setlength{\leftmargin}{2em} \setlength{\labelwidth}{1.5em}
      \setlength{\labelsep}{0.5em} } }

\newcommand{\squishend}{
    \end{list}  }

{}
\newtheorem{thm}{Theorem}{}
{}
{}
{}

\newcommand{\real}{\mbox{$\mathbb{R}$}}

\newcommand{\gauss}{\mbox{${\cal N}$}}

\newcommand{\myvec}[1]{\mbox{$\mathbf{#1}$}}
\newcommand{\myvecsym}[1]{\mbox{$\boldsymbol{#1}$}}

\newcommand{\vmu}{\mbox{$\myvecsym{\mu}$}}

\newcommand{\vphi}{\mbox{$\myvecsym{\phi}$}}

\newcommand{\vpsi}{\myvecsym{\psi}}

\newcommand{\vg}{\mbox{$\myvec{g}$}}

\newcommand{\vm}{\mbox{$\myvec{m}$}}

\newcommand{\vs}{\mbox{$\myvec{s}$}}

\newcommand{\vu}{\mbox{$\myvec{u}$}}

\newcommand{\vw}{\mbox{$\myvec{w}$}}

\newcommand{\vx}{\mbox{$\myvec{x}$}}

\newcommand{\vy}{\mbox{$\myvec{y}$}}

\newcommand{\vA}{\mbox{$\myvec{A}$}}

\newcommand{\vC}{\mbox{$\myvec{C}$}}
\newcommand{\vD}{\mbox{$\myvec{D}$}}

\newcommand{\vG}{\mbox{$\myvec{G}$}}
\newcommand{\vH}{\mbox{$\myvec{H}$}}
\newcommand{\vI}{\mbox{$\myvec{I}$}}

\newcommand{\vK}{\mbox{$\myvec{K}$}}

\newcommand{\vM}{\mbox{$\myvec{M}$}}
\newcommand{\vN}{\mbox{$\myvec{N}$}}

\newcommand{\vS}{\mbox{$\myvec{S}$}}

\newcommand{\vU}{\mbox{$\myvec{U}$}}

\newcommand{\vX}{\mbox{$\myvec{X}$}}

\newcommand{\be}{\begin{equation}}
\newcommand{\ee}{\end{equation}}
\newcommand{\bea}{\begin{eqnarray}}
\newcommand{\eea}{\end{eqnarray}}
\newcommand{\beaa}{\begin{eqnarray*}}
\newcommand{\eeaa}{\end{eqnarray*}}

\newcommand{\expm}{\mathrm{Expm}}

\usepackage{tikz}
\usepackage{tikzscale}

\usetikzlibrary{external}
\usetikzlibrary{cd}
\tikzcdset{
  arrow style=tikz,
  diagrams={>={Straight Barb}}
}

\usetikzlibrary{patterns,positioning,arrows}

\usepackage{pgfplots}
\DeclareUnicodeCharacter{2212}{−}
\usepgfplotslibrary{groupplots,dateplot}
\usetikzlibrary{shapes.arrows,shapes.misc}
\pgfplotsset{compat=newest}
\usepgfplotslibrary{groupplots}

\pgfplotsset{compat=1.11,
  /pgfplots/ybar legend/.style={
    /pgfplots/legend image code/.code={%
      \draw[##1,/tikz/.cd,yshift=-0.25em]
      (0cm,0cm) rectangle (3pt,0.8em);},
  },
}

\usepackage{tabularx} %

\begin{document}

\newcommand{\papertitle}{Structured Inverse-Free Natural Gradient Descent:\\ Memory-Efficient \& Numerically-Stable KFAC}
\newcommand{\papertitleshort}{Structured Inverse-Free Natural Gradient Descent (SINGD)}

\icmltitlerunning{\papertitleshort}
\twocolumn[
\icmltitle{\papertitle}

\icmlsetsymbol{equal}{*}

\begin{icmlauthorlist}
\parbox{\linewidth}{%
\centering
  \icmlauthor{Wu Lin}{equal,vector}
  \icmlauthor{Felix Dangel}{equal,vector}
  \icmlauthor{Runa Eschenhagen}{cambridge}
  \icmlauthor{Kirill Neklyudov}{vector}
  \icmlauthor{Agustinus Kristiadi}{vector}\\
  \icmlauthor{Richard E. Turner}{cambridge}
  \icmlauthor{Alireza Makhzani}{vector,uoft}
  }%
\end{icmlauthorlist}

\icmlaffiliation{vector}{Vector Institute}
\icmlaffiliation{cambridge}{University of Cambridge}
\icmlaffiliation{uoft}{University of Toronto}

\icmlcorrespondingauthor{Wu Lin}{yorker.lin@gmail.com}

\icmlkeywords{Machine Learning, ICML}

\vskip 0.3in
]

\printAffiliationsAndNotice{\icmlEqualContribution} %

\begin{abstract}
  Second-order methods such as KFAC can be useful for neural net training.
  However, they are often memory-inefficient since their preconditioning Kronecker factors are dense, and numerically unstable in low precision as they require matrix inversion or decomposition.
  These limitations render such methods unpopular for modern mixed-precision training.
  We address them by (i) formulating an \emph{inverse-free} KFAC update and (ii) imposing \emph{structures} in the Kronecker factors, resulting in \emph{structured inverse-free natural gradient descent (SINGD)}.
  On modern neural networks, we show that SINGD is memory-efficient and numerically robust, in contrast to KFAC, and often outperforms AdamW even in half precision.
  Our work closes a gap between first- and second-order methods in modern low-precision training.
\end{abstract}

\section{Introduction}
\label{sec:intro}

The continuing success of deep learning (DL) is---to a large extent---powered by scaling up computational power~\citep{thompson2020computational} to increase the number of trainable neural network (NN) parameters.
Contemporary natural language processing \citep{radford2019gpt2,brown2020gpt3,touvron2023llama} and computer vision \citep{dehghani2023scaling} models often consist of billions of parameters, and will likely grow further in the future.
To compensate for increasing computational demands, many training pipelines use lower precision data types~\citep{micikevicius2018mixed} and memory-efficient first-order optimizers like SGD~\citep{robbins1951stochastic} or Adam(W)~\citep{kingma2014adam,loshchilov2019adamw}.

Second-order methods, like natural gradient descent~\citep[NGD,][]{amari1998natural}, leverage curvature information which has many applications in DL:
It is useful for improving training dynamics~\citep{martens2015optimizing,osawa2023pipefisher}, pruning~\citep{wang2019eigendamage}, understanding the influence of training examples~\citep{bae2022if}, and uncertainty estimation~\citep{zhang2018noisy,immer2021scalable,daxberger2021laplace}.
One major obstacle why those methods are rarely used is their higher memory consumption and iteration cost.

The perhaps most common concept to scale second-order methods for DL is Kronecker-factored approximate curvature~\citep[KFAC,][]{heskes2000natural,martens2015optimizing} which approximates the Fisher's block diagonals via Kronecker products.
The KFAC optimizer built on top of this curvature approximation, and its variants such as \citet{george2018fast} show promising results for medium-sized NNs~\citep[e.g.][]{osawa2023pipefisher}, its usefulness is often limited by (i) memory consumption, and (ii) the use of low-precision floating-point (FP) training that renders
matrix decompositions/inversions required to pre-condition the gradient numerically unstable.

Recently, \citet{lin2023simplifying} proposed an inverse-free Kronecker-factored natural gradient descent (INGD) algorithm that replaces matrix inversion with subtraction in a matrix logarithm space.
Their update is purely based on matrix multiplications and therefore numerically stable in single-precision (FP-32); however, it is unclear whether this extends to half-precision (BFP-16).
Furthermore, INGD has not been derived from the popular natural gradient approaches for DL.
It is unclear if and how the method is connected to the predominant KFAC optimizer.
Also, INGD does not improve over KFAC's memory complexity since its Kronecker factors are dense matrices of the same size.
And lastly, INGD has only been tested on convolution-based models and it is unclear whether it is useful for training modern transformer-based architectures~\citep{vaswani2017attention}.

\begin{figure*}[!t]
  \centering
  \includegraphics[width=\linewidth]{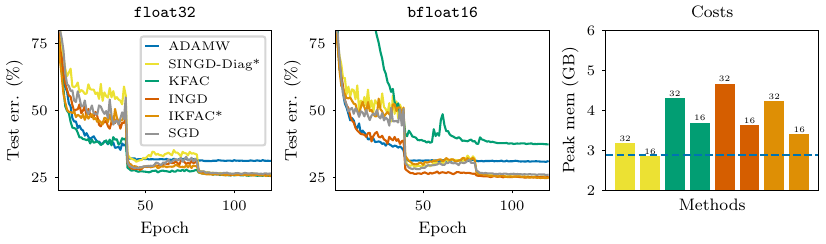}
  \caption{ CIFAR-100 experiments on VGG net.
    \emph{Left/Center:} Our methods (IKFAC and SINGD) outperform AdamW and perform stably in FP-32 \emph{and} BFP-16---unlike KFAC---as they do not require matrix inversions.
    IKFAC effectively performs KFAC updates and achieves similar performance in FP-32.
    For this task, replacing the dense Kronecker factors (INGD = SINGD-Dense) with diagonal ones (SINGD-Diag) does not harm performance while reducing cost.
    \emph{Right:} Memory consumption.
    Removing Riemannian momentum (IKFAC) or using structured Kronecker factors (SINGD-Diag) reduces INGD's memory in FP-32 and BFP-16.
    In BFP-16, SINGD-Diag achieves AdamW's memory consumption (dashed line).
  }\label{fig:one}
\end{figure*}

Here, we extend INGD to lower its computational cost and theoretically resolve its connection to other approximate NGD methods for DL (overview in \Cref{fig:overview_methods}): First, we show that a special case of INGD recovers the KFAC method.
This allows us to effectively perform KFAC updates in an \emph{inverse-free} fashion.
We call this modification of INGD \emph{inverse-free KFAC (IKFAC)}.
Second, we exploit an algebraic structure in the matrix logarithm space and propose structure-preserving updates to maintain sparse structures on Kronecker factors.
This significantly reduces memory and leads to a novel, scalable second-order optimization algorithm we call \emph{structured inverse-free natural gradient descent (SINGD)} which contains INGD and IKFAC as special cases.
We evaluate SINGD on convolution- and transformer-based models and show that it can (i) outperform SGD and AdamW while using as little memory as the latter thanks to structured Kronecker factors and (ii) yield better performance than KFAC while being stable in half-precision:
\begin{enumerate}[(a)]
\item We bridge the gap between INGD \citep{lin2023simplifying} and the original KFAC \citep{martens2015optimizing}, whose matrix inversions are unstable in low precision.
  Thereby, we effectively make KFAC inverse-free and amenable to low-precision training (\Cref{fig:one}, \emph{left/center}).
\item We impose various structures (block-diagonal, low-rank, Toeplitz, hierarchical) on INGD's Kronecker factors, allowing them to be sparse to lower the memory consumption and run time (\Cref{fig:one}, \emph{right} and \Cref{tab:performance-cifar-100}).
  Unlike many existing second-order methods tailored to a form of structure, our proposed update rule (\Cref{fig:matDL_opt}) is unified, efficient, and inverse-free for a range of structures.
  We analyze the impact of structures on downstream performance and find that structures with considerably lower memory consumption (even lower than AdamW) can yield competitive performance.
\item Unlike other second-order methods, we show that SINGD can stably train a range of modern architectures (transformers, CNNs, GNNs) in BFP-16.
  In contrast to first-order methods which are often useful in narrower scopes (SGD is best for CNNs, AdamW is best for transformers), SINGD works well and outperforms SGD and AdamW in many cases (see~\Cref{sec:experiment}).
\end{enumerate}
Our work closes a gap between first- and second-order methods in modern low precision neural network training\footnote{PyTorch implementation: \href{https://github.com/f-dangel/singd}{\texttt{github.com/f-dangel/singd}}}.

\begin{table}[!t]
  \centering
  \caption{Training times and memory consumption for the optimizers shown in \Cref{fig:one} (parenthesized values are normalized relative to SGD; our methods are marked with an asterisk).
    INGD has 80\,\% time and 30\,\% memory overhead compared to SGD.
    In contrast, our SINGD-Diag only has 30\,\% time and 2\,\% memory overhead.
    This means that by using structures we can reduce INGD's time overhead by more than half, and basically eliminate its memory overhead compared to first-order competitors.}
  \label{tab:performance-cifar-100}
  \begin{small}
    \begin{tabular}{lcc}
      \toprule
      \multirow{2}{*}{\textbf{Method}} & \textbf{Peak memory} & \textbf{Training time}
      \\
                                       & \textbf{[GiB]} & \textbf{[min]}
      \\
      \midrule
      SGD (BFP-16) & 2.63 (1.00\,x) & 18.5 (1.00\,x)
      \\
      AdamW (BFP-16) & 2.69 (1.02\,x) & 19.7 (1.07\,x)
      \\
      SINGD-Diag* (BFP-16) & 2.67 (1.02\,x) & 23.8 (1.29\,x)
      \\
      IKFAC* (BFP-16) & 3.18 (1.21\,x) & 34.0 (1.84\,x)
      \\
      INGD (BFP-16) & 3.39 (1.29\,x) & 34.1 (1.84\,x)
      \\
      KFAC (FP-32) & 4.00 (1.52\,x) & 83.2 (4.49\,x)
      \\
      \bottomrule
    \end{tabular}
  \end{small}
\end{table}

\section{Preliminaries}
\label{sec:prelim}

We first introduce the necessary ingredients to establish a connection between INGD and KFAC, which are derived from different perspectives.
We start by describing Newton's method since both methods can be seen as approximate Newton methods using NGD.
NN training often corresponds to an unconstrained minimization problem.
Consider training a NN for image classification.
Given a set of $N$ examples $\{y_i,\vx_i\}_{i=1}^{N}$ with labels $y_i$ and images $\vx_i$, the optimization problem is
\begin{equation}
  \min_{\mu } \ell(\vmu; \vy,\vX)\coloneq \min_{\mu } \textstyle\sum_{i=1}^{N} c(y_i, f(\vmu; \vx_i))\,,  \label{eq:opt_obj}
\end{equation} where $\vy\coloneq(y_1,\dots,y_N)$,
$\vX\coloneq(\vx_1,\dots,\vx_N)$, and $\hat{y}_i\coloneq f(\vmu;\vx_i)$ is a NN that outputs a predicted label $\hat{y}_i$ for an image $\vx_i$.
Parameters $\vmu$ denote learnable weights of the NN and $c(y_i,\hat{y}_i)$ is a differentiable loss function to measure the difference between a true label $y_i$ and a predicted label $\hat{y}_i$.
To solve \Cref{eq:opt_obj}, Newton's method follows the update
\begin{equation}
  \vmu \leftarrow \vmu -  \vS^{-1} \left(  \nabla_{\mu}  \ell(\vmu; \vy, \vX)    \right)\,, \label{eq:newton}
\end{equation}
where $\vS: = \nabla_{\mu}^2 \ell(\vmu; \vy, \vX)$ is the Hessian of the loss.

\subsection{KFAC: Approximate NGD for MLE}
\label{subsec:kfac}

\begin{figure*}[t]
  \centering
  \begin{minipage}[b]{0.61\linewidth}
    \tikzexternaldisable
    \resizebox{\linewidth}{!}{%
      \includegraphics{figs/overview_methods}
    }
    \tikzexternalenable
  \end{minipage}
  \hfill
  \begin{minipage}[b]{0.38\linewidth}
    \caption{ Existing methods and their relation to our proposed methods.
      IKFAC behaves like KFAC (\Cref{thm:kfac_K_part}), but is numerically stable in low precision.
      In contrast to IKFAC, INGD has Riemannian momenta and adaptive damping and curvature, which can yield better performance in practice (\Cref{sec:experiment}).
      INGD is equivalent to SINGD with unstructured Kronecker factors (SINGD-Dense).
      Structured Kronecker factors reduce memory and computational cost.
    }
    \vspace{-3.5ex}
    \label{fig:overview_methods}
  \end{minipage}
\end{figure*}

Computing the Hessian, as required by Newton's method,
 is usually intractable for NNs.
NGD uses a Fisher information matrix (FIM) instead of the Hessian by reformulating problem \eqref{eq:opt_obj} as maximum likelihood estimation (MLE) of $p(\vy \mid \vmu,\vX)=\prod_{i}p(y_i \mid \vmu,\vx_i)$, where $p(y_i \mid \vmu,\vx_i)\coloneq \exp(-c(y_i, f(\vmu,\vx_i)) )$.
The maximization problem $\max_{\mu } p(\vy\mid\vmu,\vX)$ is equivalent to the MLE problem
\begin{equation}
  \min_{\mu} -\log p(\vy\mid \vmu,\vX) = \min_{\mu} \ell(\vmu; \vy,\vX)\,.
  \label{eq:mle_opt}
\end{equation}
This formulation allows to exploit additional statistical structures such as the FIM which is defined as shown below \citep{kunstner2019limitations}, where we assume a label $y$ is sampled from the likelihood $p(y\mid\vmu,\vx_i)$ given an image $\vx_i$.
With $\vs_i(y) \coloneq \log p(y \mid \mu, \vx_i)$, we have
\begin{align}\label{eq:FIM}
  \begin{split}
    \fim(\vmu)
    &\coloneq \sum_{i=1}^{N} \mathbb{E}_{y \sim p(y \mid \mu, x_i)}\left[  \nabla_{\mu} \vs_i(y) (\nabla_{\mu} \vs_i(y))^\top   \right]
    \\
    &= \sum_{i=1}^{N}  \mathbb{E}_{y \sim p(y \mid \mu, x_i)}\left[ -\nabla_{\mu}^2 \vs_i(y) \right]\,.%
  \end{split}
\end{align}
For ubiquitous loss functions like the mean-squared error and cross-entropy, and more generally, many members of the exponential family with natural parameterization, the FIM coincides with the generalized Gauss-Newton (GGN) matrix \citep{wang2010fisher,martens2014new}, a common approximation of the Hessian in deep learning \citep{schraudolph2002fast,botev2017practical}.
This relationship connects NGD to Newton's method.
A common approximation of the FIM/GGN and Hessian is the so-called \emph{empirical} Fisher $\smash{\hat{\fim}} (\vmu)$, which replaces the samples $y$ from the model's predictive distribution in \cref{eq:FIM} with the empirical data labels $y_i$:
\begin{align*}
  \begin{split}
    \hat{\fim} (\vmu)
    &\coloneq
      \sum_{i=1}^{N} \nabla_{\mu} \vs_i(y_i) (\nabla_{\mu} \vs_i(y_i))^\top
    \\
    &\approx
      -  \sum_{i=1}^{N} \nabla_{\mu}^2 \vs_i(y_i) = \vS\,.
  \end{split}
\end{align*}
While there is no clear theoretical justification for this Hessian approximation \citep{kunstner2019limitations}, it simplifies the implementation, reduces cost, and has been shown to work well in practice \citep{graves2011practical,osawa2019practical}.
This approximation is also known as Fisher's scoring with observed FIM for nonlinear models \citep{osborne1992fisher,smyth1996partitioned,smyth2015optimization}.
With this, we can formulate an NGD update with the \emph{empirical} FIM $\smash{\hat{\fim}} (\vmu)$ to approximate Newton's method as
\begin{align*}
  \vmu
  &\leftarrow
    \vmu -  \stepsize   \left( \hat{\fim} (\vmu) \right)^{-1} \nabla_{\mu} \ell(\vmu;\vy,\vX)
  \\
  &\phantom{\leftarrow} \approx \vmu -  \stepsize    \vS^{-1} \nabla_{\mu} \ell(\vmu;\vy,\vX).
\end{align*}
We call this update NGD for MLE.

KFAC \cite{heskes2000natural,martens2015optimizing} is the probably most common second-order optimizer in DL.
The KFAC algorithm is based on a Kronecker-factored approximation of the Fisher, which is also sometimes referred to as KFAC.
Here, we refer to the algorithm as \emph{KFAC} or \emph{KFAC method} and to the approximation as \emph{Kronecker approximation}; we will consider the \emph{empirical} Fisher's Kronecker approximation.
It approximates the per-layer FIM with a Kronecker-factored block $\smash{\tilde{\fim}}_{l}$ for each layer $l$ of the net.
This approximation has first been derived for linear layers, later for convolutional \citep{grosse2016kronecker} and recurrent layers \citep{martens2018kroneckerfactored}, and recently been generalized to all linear layers that use weight sharing \citep{eschenhagen2023kronecker}, e.g.\,graph neural networks and transformers.
A block is given by $\smash{\tilde{\fim}_{l}} (\vmu) \coloneq \vU_{l} \otimes \vG_{l},$ with $\vU_{l} \coloneq \vu_{l} \smash{\vu_{l}^\top} \in \real^{d_i \times d_i}$ and $\vG_{l} \coloneq \vg_{l} \smash{\vg_{l}^\top} \in \smash{\real^{d_o \times d_o}}$,
where $\vu_{l} \in \smash{\real^{d_i}}$ is the $l$th layer's input and $\vg_{l} \in \smash{\real^{d_o}}$ is the gradient of the loss w.r.t.\,the layer's output.
We suppress the dependence on the parameters $\vmu$ and the input $\vx_i$ and, for simplicity, assume no weight sharing.
KFAC also uses exponential moving averages ($\beta_1$) over $\vU$ and $\vG$ (yielding $\vS_{K}, \vS_{C}$) and damping $\lambda$, see \cref{fig:matDL_opt2}.

While the Kronecker approximation enables more efficient gradient preconditioning, KFAC needs to store the dense Kronecker factors $\vS_K$ and $\vS_C$ and invert them at every preconditioner update.
The run time overhead is usually amortized by updating the preconditioner less frequently, but this can cause instabilities, especially in low-precision settings.
Second, the Kronecker factors introduce significant memory overhead, which poses issues in large models.
Since low-precision training is becoming the standard norm in fields like natural language processing, these issues will become more apparent in modern DL.
There are multiple numerical concerns when using KFAC or variants thereof in low precision.
In PyTorch \citep{paszke2019pytorch} and JAX \citep{jax2018github} implementations, all tensors must be casted into FP-32 as (B)FP-16 matrix inverses/decompositions are not supported.
Moreover, $\vg_l$ has to be rescaled to avoid over- or under-flows when calculating $\vG_l$.
Memory consumption has previously been addressed through diagonal or block-diagonal versions of $\vU_{l}, \vG_{l}$ \citep{zhang2018noisy,grosse2023studying}.
However, it is unclear if these simple structures maintain downstream performance.

\subsection{INGD: Approximate NGD for Bayesian estimation}
\label{subsec:localcov}

Derived from Bayesian principles, INGD \citep{lin2023simplifying} directly approximates the Hessian inverse.
We first introduce two ingredients INGD builds on: the Bayesian learning rule \citep[BLR,][]{khan2017conjugate,zhang2018noisy,khan18a,osawa2019practical,lin2020handling,khan2021bayesian,tan2022analytic} and an inverse-free second-order method from \citet{lin2021tractable}.
By the BLR, Newton's method to solve the MLE \eqref{eq:mle_opt} can be seen as another natural-gradient update to solve a variational inference (VI) problem with a delta approximation \citep{khan2021bayesian}.
This interpretation allows to view a precision matrix in the variational problem as Hessian estimation in the MLE problem.
Thus, \citet{lin2021tractable} suggest reparameterizing the Hessian as the precision of the Gaussian posterior in a matrix logarithm space and exploiting the parameterization invariance of natural gradients to obtain an inverse-free update.

\paragraph{BLR} Consider a Bayesian problem formulation, where NN weights are random variables.
We denote these weights by new parameters $\vw$ since random variables are no longer learnable and use a variational Gaussian distribution to approximate the posterior over the random variables.
Its mean and precision will be treated as the learnable weights $\vmu$ and the Hessian estimation $\vS$ in Newton's step \eqref{eq:newton}.

The VI problem considered in the learning rule is defined as $\min_{\sgparam   }  - \elbofinal(\gparam)$ with the evidence lower bound (ELBO)
\begin{align}
  \begin{split}
    \elbofinal(\gparam)
    &\coloneq
      \mathbb{E}_{w \sim q(w \mid \sgparam)} \left[ \log p(\vw) + \log p(\vy \mid \vw,\vX) \right]
    \\
    &\phantom{\coloneq} + H_q(\gparam)\,. \label{eq:blr_opt}
  \end{split}
\end{align}
$\gparam=\{\vmu,\vS\}$ are the learnable parameters of the variational Gaussian distribution $q(\vw\mid\gparam)=\gauss(\vw\mid\vmu,\vS)$ with mean $\vmu$ and precision $\vS$.
The likelihood $p(\vy\mid\vw,\vX)=\exp( - \ell(\vw ; \vy,\vX))$ takes the same form as in the MLE setting while the prior $p(\vw)\propto \exp(-R(\vw))$ is defined by a regularizer $R(\vw) \geq 0$.
To recover the MLE problem, we consider an uninformative prior $p(\vw)$ (i.e., $R(\vw)=0$).
$H_q(\gparam)\coloneq \mathbb{E}_{w \sim q} \left[ - \log q \right] $ is the entropy of $q(\vw \mid \gparam)$.

Similar to the MLE case, the Bayesian formulation allows to exploit additional statistical structures in form of another FIM, which is that of the variational Gaussian defined as
\begin{align*}
  \fim(\gparam)
  &\coloneq
    \mathbb{E}_{w \sim q(w \mid \sgparam)}\left[  \nabla_{\sgparam} \log q(\vw \mid\gparam) \nabla_{\sgparam}^\top \log q(\vw\mid\gparam)    \right]
  \\
  &= - \mathbb{E}_{w \sim q}\left[  \nabla_{\sgparam}^2 \log q(\vw \mid \gparam)  \right]\,,
\end{align*}
and has a closed-form expression. This FIM should \emph{not} be confused with the FIM used for MLE \eqref{eq:FIM}.

Under the BLR, we perform NGD updates not only on $\vmu$ but also on $\vS$.
\citet{khan2021bayesian} formulate a step with the \emph{exact} FIM $\fim(\gparam)$  and stepsize $\stepsize>0$ to update $\gparam=\{\vmu,\vS\}$,
\begin{equation*}
  \gparam \leftarrow  \gparam - \stepsize  \Big(\fim(\gparam)\Big)^{-1} \nabla_\sgparam \left( - \elbofinal(\gparam) \right)\,.
\end{equation*}
This is the NGD update for BLR, vis-\`{a}-vis for MLE.
Following \citet{khan2018fast}, the update simplifies to
\begin{align*}
  \vS &\leftarrow   (1-\stepsize) \vS + \stepsize   { \color{red} \mathbb{E}_{w \sim q(w \mid \mu,S)} \left[ \nabla_{w}^2 \ell(\vw ; \vy,\vX) \right] }\,,
  \\
  \vmu &\leftarrow   \vmu - \stepsize \vS^{-1} { \color{red} \mathbb{E}_{w \sim q(w \mid \mu,S)} \left[   \nabla_{w}  \ell(\vw ; \vy,\vX)   \right]  }\,.
\end{align*}
Further simplifying expectations with a delta approximation (highlighted in red) at mean $\vmu$, we obtain
\begin{align*}
  \vS &\leftarrow  (1-\stepsize) \vS + \stepsize   { \color{red}  \nabla_{\mu}^2 \ell(\vmu ; \vy,\vX)}\,,
  \\
  \vmu &\leftarrow   \vmu - \stepsize \vS^{-1}  { \color{red}  \nabla_{\mu}  \ell(\vmu ; \vy,\vX)  }\,.
\end{align*}
which recovers Newton's method in \eqref{eq:newton} for $\stepsize=1$.

\begin{figure*}[!t]
  \centering
  \begin{minipage}{.48\textwidth}
    \textbf{ KFAC
      (\citeauthor{martens2015optimizing}, \citeyear{martens2015optimizing})
    }

    \begin{algorithmic}[1]
      \STATE
      \footnotesize  Each $T$ iters, update   \scalebox{1.0}{  $\vS_K  $, $\vS_C  $}   \\
      Obtain  $ \vU \otimes \vG $ to approximate $\nabla_\mu^2 \ell(\vmu)$ \\
      \scalebox{1.0}{   $ \vS_K \leftarrow (1-\stepsize_1) \vS_K + \stepsize_1 \vU $} \\
      \scalebox{1.0}{  $\vS_C \leftarrow (1-\stepsize_1) \vS_C + \stepsize_1 \vG  $} \\
      \scalebox{1.0}{  $\vS_K^{-1} \leftarrow \left( \vS_K  + \lambda \vI_{d_i} \right)^{-1} $} \\
      \scalebox{1.0}{  $\vS_C^{-1} \leftarrow \left( \vS_C  + \lambda \vI_{d_o}  \right)^{-1} $} \\

      \STATE
      \scalebox{1.0}{    $\vm_\mu  \leftarrow \alpha_2 \vm_\mu  + \vS_C^{-1} \mathrm{vec}^{-1}( \vg )\vS_K^{-1}  + \gamma \mathrm{vec}^{-1}(\vmu) $}
      \STATE
      \scalebox{1.0}{   $ \vmu \leftarrow \vmu - \stepsize_2 \mathrm{vec}(\vm_\mu )$}
    \end{algorithmic}
  \end{minipage}
  \begin{minipage}{.48\textwidth}
    \textbf{IKFAC (ours)}
    \begin{algorithmic}[1]
      \STATE
      \footnotesize   Each $T$ iters, update   \scalebox{1.0}{ $\vm_K$, $\vm_C$, $\vK$, $\vC$}   \\
      Obtain  $ \vU \otimes \vG $ to approximate $\nabla_\mu^2 \ell(\vmu)$ \\
      \scalebox{1.0}{   $\vm_K  \leftarrow {\color{red} 0 }\vm_K + \frac{1}{2 d_o}( {\color{red} d_o} \vH_K + {\color{red}  \lambda d_o } \vK^\top\vK- d_o\vI_{d_i} )$} \\
      \scalebox{1.0}{  $\vm_C  \leftarrow {\color{red} 0}  \vm_C + \frac{1}{2 d_i}( {\color{red} d_i } \vH_C +  {\color{red}   \lambda d_i} \vC^\top\vC- d_i\vI_{d_o} )$} \\
      \scalebox{1.0}{  $\vK \leftarrow  \vK (\vI_{d_i}-\stepsize_1\vm_K)$} \\
      \scalebox{1.0}{  $\vC \leftarrow  \vC (\vI_{d_o}-\stepsize_1\vm_C)$} \\

      \STATE
      \scalebox{1.0}{    $\vm_\mu  \leftarrow \alpha_2 \vm_\mu  + \vC\vC^{\top} \mathrm{vec}^{-1}( \vg) \vK\vK^{\top}  + \gamma \mathrm{vec}^{-1}(\vmu) $}
      \STATE
      \scalebox{1.0}{   $ \vmu \leftarrow \vmu - \stepsize_2 \mathrm{vec}(\vm_\mu )$}
    \end{algorithmic}
  \end{minipage}
  \caption{ Comparison between KFAC and IKFAC update for one weight matrix $\mathrm{vec}^{-1}(\vmu) \in \real^{d_o \times d_i}$.
    The flattened gradient is $\vg\coloneq\nabla_\mu \ell(\vmu) \in \real^{d_o d_i}$ and $\mathrm{vec}^{-1}(\vg) \in \real^{d_o \times d_i}$ is its matrix reshape.
    IKFAC uses $\vH_K \coloneq \vK^\top \vU \vK$ and $\vH_C \coloneq \vC^\top \vG \vC$ to incorporate the Kronecker curvature $\vU$ and $\vG$.
    Both methods use momentum buffers $\vm_\mu$ for the weight-decayed update direction with momentum $\alpha_2$ and weight decay $\gamma$, and a learning rate $\beta_2$ for the parameter update.
    (\emph{Left}) KFAC uses an exponentially moving average with decay $1 - \beta_1$ to accumulate the Kronecker factors and applies a damping term $\lambda \vI$ before inversion to handle potential singularities in $ \vS_K $, $ \vS_C $.
    (\emph{Right}) In contrast to KFAC, IKFAC directly approximates $\smash{(\vS_K +\lambda \vI)^{-1}}$ and $\smash{(\vS_C +\lambda \vI)^{-1}}$ by $\vK \smash{\vK^\top}$ and $\vC \smash{\vC^\top}$.
    The pre-conditioner update is a modification of INGD~\citep{lin2023simplifying} and the changes---zero Riemannian momentum, and non-adaptive damping and curvature---are highlighted in \textcolor{red}{red}.
  }
  \label{fig:matDL_opt2}

  \vspace{1.5ex}

  \newcommand{\highlightAlt}[1]{\textcolor{red}{#1}}
  \newcommand{\highlight}[1]{\textcolor{blue}{#1}}
  \newcommand{\structured}[1]{\ensuremath\highlight{\hat{\mathcal{L}}}_{#1}}
  \newcommand{\project}[2]{\ensuremath\highlight{\hat{\Pi}}_{#1}\highlight{(}#2\highlight{)}}
  \center
  \begin{minipage}[t]{.48\textwidth}
    \textbf{ INGD (\citeauthor{lin2023simplifying}, \citeyear{lin2023simplifying}) %
    }

    \begin{algorithmic}[1]
      \STATE
      \footnotesize   Each $T$ iterations, update   \scalebox{1.0}{ $\vm_K$, $\vm_C$, $\vK$, $\vC$}   \\
      Obtain  $ \vU \otimes \vG $ to approximate $\nabla_\mu^2 \ell(\vmu)$ \\[0.2ex]
      \scalebox{0.8}{   $\vm_K  \leftarrow { \color{red} \alpha_1} \vm_K + \frac{1}{2 d_o}(
        { \color{red} \mathrm{Tr}(\vH_C) } \vH_K + {\color{red} c^2} \vK^\top\vK - d_o\vI_{d_i} )$}
      \\
      \scalebox{0.8}{  $\vm_C  \leftarrow { \color{red} \alpha_1} \vm_C + \frac{1}{2 d_i}({ \color{red} \mathrm{Tr}(\vH_K) } \vH_C +  { \color{red} \kappa^2 } \vC^\top\vC- d_i\vI_{d_o} )$}
      \\
      \scalebox{1.0}{  $\vK \leftarrow  \vK (\vI_{d_i}-\stepsize_1\vm_K)$} \\
      \scalebox{1.0}{  $\vC \leftarrow  \vC (\vI_{d_o}-\stepsize_1\vm_C)$} \\

      \STATE
      \scalebox{1.0}{    $\vm_\mu  \leftarrow \alpha_2 \vm_\mu  + \vC\vC^{\top} \mathrm{vec}^{-1}( \vg) \vK\vK^{\top}  + \gamma  \mathrm{vec}^{-1}(\vmu) $}
      \STATE
      \scalebox{1.0}{   $ \vmu \leftarrow \vmu - \stepsize_2 \mathrm{vec}(\vm_\mu )$}
    \end{algorithmic}
  \end{minipage}
  \begin{minipage}[t]{.48\textwidth}
    \textbf{\highlight{S}INGD (ours)}

    \begin{algorithmic}[1]
      \STATE
      \footnotesize   Each $T$ iterations, update   \scalebox{1.0}{ $\structured{m_K}$, $\structured{m_C}$, $\structured{K}$, $\structured{C}$}   \\
      Obtain  $ \vU \otimes \vG $ to approximate $\nabla_\mu^2 \ell(\vmu)$ \\
      \scalebox{0.8}{   $\structured{m_K}  \leftarrow\highlightAlt{\alpha_1} \structured{m_K} + \frac{1}{2 d_o}\project{K}{
          \highlightAlt{\mathrm{Tr}(\vH_{\structured{C}})}
          \vH_{\structured{K}} +  \highlightAlt{c^2} (\structured{K})^\top\structured{K}- d_o \vI_{d_i} }$} \\
      \scalebox{0.8}{  $\structured{m_C}  \leftarrow
        \highlightAlt{\alpha_1}
        \structured{\vm_C} + \frac{1}{2 d_i}\project{C}{
          \highlightAlt{
            \mathrm{Tr}(\vH_{\structured{K}})
          }
          \vH_{\structured{C}} +  \highlightAlt{\kappa^2} (\structured{C})^\top\structured{C}- d_i \vI_{d_o} }$} \\
      \scalebox{1.0}{  $\structured{K} \leftarrow \structured{K} (\vI_{d_i}-\stepsize_1\structured{m_K})$} \\
      \scalebox{1.0}{  $\structured{C} \leftarrow \structured{C} (\vI_{d_o}-\stepsize_1\structured{m_C})$} \\

      \STATE
      \scalebox{0.9}{    $\vm_\mu  \leftarrow \alpha_2 \vm_\mu  +
        \structured{C}(\structured{C})^{\top} \mathrm{vec}^{-1}( \vg) \structured{K}(\structured{K})^{\top}  + \gamma \mathrm{vec}^{-1}(\vmu) $}
      \STATE
      \scalebox{1.0}{   $ \vmu \leftarrow \vmu - \stepsize_2 \mathrm{vec}(\vm_\mu ) $}
    \end{algorithmic}
  \end{minipage}

  \caption{
    Comparison of a single weight matrix's update between INGD and our extension---SINGD---via structured Kronecker factors.
    (\emph{Left}) INGD features \highlightAlt{Riemannian momentum} ($\alpha_1$), \highlightAlt{adaptive curvature} ($\mathrm{Tr}(\vH_C)$, $\mathrm{Tr}(\vH_K)$),  \highlightAlt{adaptive damping} ($c^2  \coloneq \lambda\mathrm{Tr}(\vC^\top\vC)$, $\kappa^2  \coloneq \lambda\mathrm{Tr}(\vK^\top\vK)\,$), and  \highlightAlt{correlated updates} of $\vK$ and $\vC$ ($\vm_K$, $\vm_C$).
    The pre-conditioner matrices are updated with a learning rate $\beta_1$, and the optimizer keeps a momentum buffer on the weight-decayed update with momentum $\alpha_2$ and weight decay $\gamma$.
    The learning rate for the parameters is $\beta_2$.
    (\emph{Right})
    SINGD's update is similar but each Kronecker factor and its momentum ($\bullet$) is replaced by its \highlight{structured version} ($\smash{\structured{\bullet}}$, e.g.\,(block-)diagonal); likewise in the computation of $c^2$, $\kappa^2$, $\vH_K$, and $\vH_C$.
    When updating the momenta, their structure is preserved through a \highlight{subspace projection map} $\smash{\project{\bullet}{\cdot}}$ that restores $\smash{\structured{\bullet}}$'s structure from a dense symmetric matrix $\cdot$ (e.g.\,taking the (block) diagonal). Importantly, we can efficiently compute the extraction map without expanding its argument in dense form, which reduces memory and run time.
    The extension of IKFAC to SIKFAC is analogous.
    One of the notable elements of INGD and SINGD is that they are scale invariant to the choice of the Kronecker approximation (see \Cref{app:invariance}) as the approximation is not unique.
  }     \label{fig:matDL_opt}
\end{figure*}

\paragraph{Removing inversion} \citet{lin2021tractable} reparameterize the precision matrix $\vS$ in a matrix logarithm space and perform natural gradient updates in this space, which transforms inversion into subtraction.
One can go back directly to the original space, without explicitly inverting a matrix, via a truncated matrix exponential.
The method is inverse-free and, since NGs are parameterization invariant, Newton-like.

The first step is to express the precision matrix $\vS$ using a non-singular square matrix $\vA$ as $\vS= \smash{\vA^{-\top}\vA^{-1}}$ and perform a natural gradient step using the exact FIM in a tangent space (denoted by $\vM$) of $\vA_t$ at iteration $t$.
We then construct a new map as $\vA\coloneq\vphi(\vA_t,\vM)\coloneq\vA_t \expm(\nicefrac{1}{2} \vM)$ using both the current point $\vA_t$ and $\vM$ as input, where $\expm(\vN)=\vI + \smash{\sum_{j=1}^{\infty} \nicefrac{\vN^j }{j!}}
$ is the matrix exponential.
Observe that $\vM$ stays in a matrix logarithm space.
At each iteration $t$, we use a new matrix logarithm space associated to $\vA_t$ and generate a new origin $\vM_0=\mathbf{0}$ in this space to represent $\vA_t$ since $\vA_t \equiv \vphi(\vA_t,\mathbf{0})= \vA_t \expm(\nicefrac{1}{2} \vM_0)$.
The map $\vphi$ is a \emph{local reparameterization} map that takes not only $\vM$ but also $\vA_t$ as input.
Thanks to this map, the Fisher block is \emph{locally orthonormalized} \citep{lin2023simplifying} at origin $\vM_0$.
Since we used the origin to represent $\vA_t$ in the local coordinate $\vM$, a natural gradient step becomes a (Euclidean) gradient step in the space of $\vM$, which makes it easy to add Riemannian momentum \citep{lin2023simplifying} into the structured positive-definite matrix $\vS$.
This allows to perform updates in the logarithmic space of $\vM$ and avoid matrix inversions:
\begin{align}\label{eq:update-log-space}
  \begin{split}
    \vM
    &\leftarrow
      \vM_0 - \stepsize  \vN\,,
    \\
    \vmu
    &\leftarrow
    \vmu - \stepsize  \vA_{t+1}   \vA_{t+1}^\top   \nabla_\mu  \ell(\vmu ; \vy,\vX)\,,
  \end{split}
\end{align}
where
$\vA_{t+1} \coloneq \vphi(\vA_t,\vM)= \vA_t \expm\left( \nicefrac{1}{2} \vM \right)$ and  $\vN \coloneq \vA_t^\top  \nabla_\mu^2 \ell(\vmu ; \vy,\vX) \vA_t  - \vI$.
\Cref{eq:update-log-space} is a Newton-like update without matrix inverse.
To see that, we can reexpress the update of $\vA$  in terms of $\vS$
and use properties of the matrix exponential function,
\begin{align*}
  \vS_{t+1}
  &=
    \vA_{t+1}^{-T} \vA_{t+1}^{-1} = \vA_t^{-T} \expm\left(\stepsize \vN \right) \vA_t^{-1}
  \\
  &=
    (1-\stepsize) \vS_t  + \stepsize  \nabla_\mu^2 \ell(\vmu ; \vy,\vX)  + O(\stepsize^2).
\end{align*}
Next, we can construct a structured precision matrix $\vS$ as a structured Hessian estimation using a sparse non-singular matrix $\vA$.
As we will discuss in \Cref{sec:structures}, it is essential to update $\vM$ to preserve sparsity in $\vA$.
The space of $\vM$ as a tangent/logarithm space of $\vA$ allows us to efficiently impose sparse structures on $\vA$ without requiring the Hessian $\nabla_\mu^2 \ell(\vmu ; \vy,\vX)$ or a Hessian approximation to be sparse or structured.
This is different from another inverse-free method \citep{tan2022analytic} that considers directly performing NGD updates of $\vA$ instead of $\vM$, where $\vA$ must be restricted to a (triangular) Cholesky factor.
This does not preserve sparsity in $\vA$ unless the Hessian or its approximation admit a special structure, which is usually not the case in DL problems.

\paragraph{INGD} Our work is built on INGD (\cref{fig:matDL_opt}) where $\vA= \vK \otimes \vC $ is factorized into two Kronecker factors.
The exact FIM under this parameterization is singular due to a correlation between $\vK$ and $\vC$: the Kronecker factorization is not unique.
\citet{lin2023simplifying} propose a (non-singular) block-diagonal approximated FIM by ignoring the correlation in the original FIM and perform NGD with this block-diagonal FIM on tangent spaces of the factors.
Riemannian momentum is further introduced in the update of $\vK$ and $\vC$.
They use the Kronecker approximation discussed in \cref{subsec:kfac} to approximate the Hessian $\nabla_\mu^2 \ell(\vmu ; \vy,\vX)$ and truncate the matrix exponential to obtain a purely matrix-multiplication based update scheme.
It is unclear how INGD is related to KFAC which uses another Kronecker factorization $\vS=\vS_K \otimes \vS_C$.
INGD also remains memory-inefficient due to the use of dense Kronecker factors.
The authors only consider and evaluate it on convolution-based models in single precision.
It remains unclear whether INGD is useful to train transformer-based models, and in half-precision.

\section{Structured inverse-free NGD}
\label{sec:method}

Inspired by INGD, we propose an inverse-free KFAC update as a specific setting of INGD to address KFAC's numerical instability in low precision.
We show that this scheme effectively recovers KFAC.
We then address the memory inefficiency of  KFAC and INGD  for training transformer-based models by extending INGD with structures.

\subsection{Inverse-free KFAC Updates for Numerical Stability}
\label{subsec:ikfac}

\begin{figure*}[t]
  \begin{minipage}[t]{0.54\linewidth}
    \centering
    \captionof{table}{ Subspaces of the  logarithm space and their projection maps $\hat{\Pi}(\vM)$, where $\vM$ is a symmetric matrix. The hierarchical structure is constructed by replacing the diagonal matrix  $\vD_{22}$  in the rank-k upper-triangular structure with another rank-$k$  triangular matrix \scalebox{0.5}{$\begin{bmatrix} \vA_{22} & \mathbf{0} \\ \vA_{23} & \vA_{33} \end{bmatrix}$} for a better  approximation.  }
    \label{tab:subspace_proj}
    \resizebox{\linewidth}{!}{%
      \newcolumntype{Y}{>{\centering\arraybackslash}X}
      \newcolumntype{s}{>{\hsize=.7\hsize}Y}
      \begin{small}
        \begin{tabularx}{0.95\textwidth}{YsY}
          \toprule
          Subspace of the log (Lie-algebraic) space
          & Matrix Lie sub-group structure in $\vK$
          & Subspace projection map $ \hat{\Pi}(\vM)$
          \\
          \midrule
          \scalebox{0.77}{$ \begin{bmatrix}
                              a_{1,1} &    0       &     \hdots    &                     0 \\
                              a_{2,1} & a_{2,2} &        &            0              \\
                              \vdots &     \vdots & \ddots &    \vdots              \\
                              a_{d_i,1} & a_{d_i,2} & \ldots &  a_{d_i,d_i}
                            \end{bmatrix}$}
          & Lower-triangular (Tril.)
          &  \scalebox{0.75}{$ \begin{bmatrix}
                                 m_{1,1} &      0           & \hdots        &                    0 \\
                                 2 m_{2,1} & m_{2,2} &        &            0             \\
                                 \vdots &     \vdots & \ddots &        \vdots          \\
                                 2 m_{d_i,1} &2 m_{d_i,2} & \ldots  & m_{d_i,d_i}
                               \end{bmatrix} $}
          \\ \\
          \scalebox{0.8}{$\begin{bmatrix}
                            \mathbf{A}_{11} & \mathbf{0}  & \cdots & \mathbf{0}  \\
                            \mathbf{0}  & \mathbf{A}_{22} & \cdots & \mathbf{0}  \\
                            \vdots          & \vdots          & \ddots & \vdots          \\
                            \mathbf{0}  & \mathbf{0}  & \cdots & \mathbf{A}_{qq}
                          \end{bmatrix} $}
          &   (Block) Diagonal (block size $k$)
          &   \scalebox{0.8}{ $ \begin{bmatrix}
                                  \mathbf{M}_{11} & \mathbf{0}  & \cdots & \mathbf{0}  \\
                                  \mathbf{0}  & \mathbf{M}_{22} & \cdots & \mathbf{0}  \\
                                  \vdots          & \vdots          & \ddots & \vdots          \\
                                  \mathbf{0}  & \mathbf{0}  & \cdots & \mathbf{M}_{qq}
                                \end{bmatrix}  $}
          \\ \\
          \scalebox{0.8}{$\begin{bmatrix}
                            \mathbf{A}_{11} & \mathbf{A}_{12}  & \mathbf{A}_{13} \\
                            \mathbf{0}  & \mathbf{A}_{22}  & \mathbf{0} \\
                            \mathbf{0}  & \mathbf{A}_{32}  & \mathbf{A}_{33}
                          \end{bmatrix}$},
          \scalebox{0.8}{$\mathbf{A}_{22}$ is diag.,}
          \scalebox{0.8}{$\mathbf{A}_{11} \in \mathbb{R}^{d_2 \times d_2}$,}
          \scalebox{0.8}{$\mathbf{A}_{33} \in \mathbb{R}^{d_3 \times d_3}$}
          &  Hierarchical ($k\coloneq d_2+d_3$)
          &  \scalebox{0.8}{ $ \begin{bmatrix}
                                 \mathbf{M}_{11} & 2 \mathbf{M}_{12}  & 2 \mathbf{M}_{13} \\
                                 \mathbf{0}  & \mathrm{Diag}( \mathbf{M}_{22})  & \mathbf{0} \\
                                 \mathbf{0}  & 2 \mathbf{M}_{32}  & \mathbf{M}_{33}
                               \end{bmatrix} $}
          \\ \\
          \scalebox{0.8}{$\begin{bmatrix}
                            {\vA}_{11} & \mathbf{A}_{12}   \\
                            \mathbf{0}  & \mathbf{D}_{22}
                          \end{bmatrix}$},   \scalebox{0.8}{$\mathbf{D}_{22}$ is diag., $\vA_{11} \in \mathbb{R}^{k \times k}$}
          &  Rank-$k$ upper-triangular
          & \scalebox{0.8}{$\begin{bmatrix}
                              {\vM}_{11} & 2 \mathbf{M}_{12}   \\
                              \mathbf{0}  &  \mathrm{Diag}(\mathbf{M}_{22})
                            \end{bmatrix}$}
          \\ \\
          \scalebox{0.7}{ $\begin{bmatrix}
                             a_{0} & a_{1}   & a_{2} & \cdots & a_{(d_i-1)}   \\
                             0 & a_{0}      & a_{1} & \ddots    & \vdots  \\
                             0 & 0      & \ddots & \ddots & a_{2}  \\
                             \vdots & \ddots & \ddots & \ddots & a_{1}   \\
                             0 &   \cdots     & \ddots & 0 & a_{0}
                           \end{bmatrix}$ }
          &  Upper-triangular Toeplitz (Triu-Toepl.)
          & \scalebox{0.7}{$\begin{bmatrix}
                              b_{0} & 2 b_{1}   & 2 b_{2} & \cdots & 2 b_{ (d_i-1) }  \\
                              0 & b_{0}      & 2 b_{1} & \ddots &    \vdots     \\
                              0 & 0      & \ddots & \ddots & 2 b_{2}   \\
                              \vdots & \ddots & \ddots & \ddots & 2 b_{1} \\
                              0 & \cdots    & \cdots & 0 & b_{0}
                            \end{bmatrix}$}
            \scalebox{0.8}{$b_j\coloneq\frac{1}{ d_i -j  } \sum_{k=1}^{ d_i -j } m_{k, k+j } $}
          \\
          \bottomrule
        \end{tabularx}%
      \end{small}
    }
  \end{minipage}
  \hfill
  \begin{minipage}[t]{0.44\linewidth}
    \centering
    \vspace{-0ex}
    \begin{small}
      \begin{tabular}{ccccccc}
        &
          $\vK$
        &
          $\vK \vK^{\top}$
        &
          $\big(\vK \vK^{\top})^{-1}$
        \\
        \raisebox{0.06\linewidth}{Dense}
        & \includegraphics[width=0.15\linewidth]{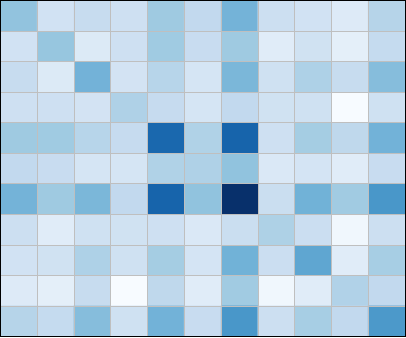}
        & \includegraphics[width=0.15\linewidth]{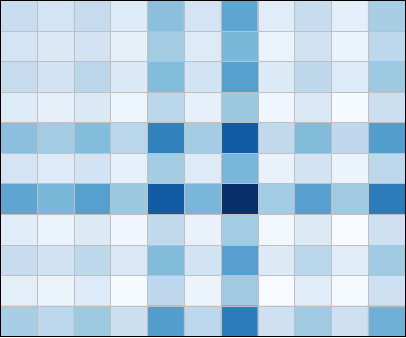}
        & \includegraphics[width=0.15\linewidth]{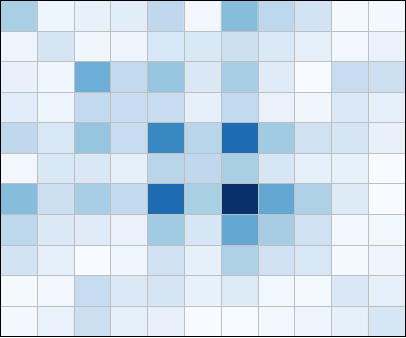}
        \\
        \raisebox{0.06\linewidth}{Diagonal}
        & \includegraphics[width=0.15\linewidth]{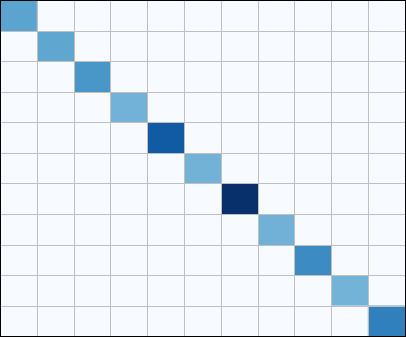}
        & \includegraphics[width=0.15\linewidth]{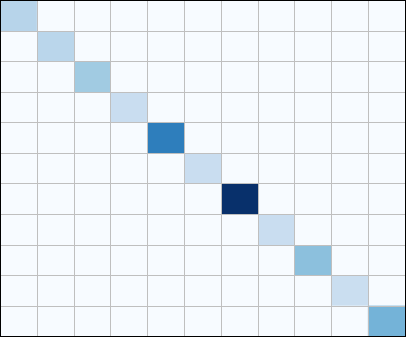}
        & \includegraphics[width=0.15\linewidth]{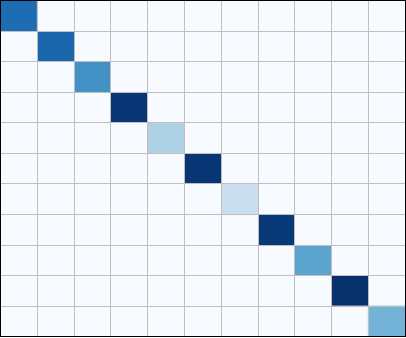}
        \\
        \raisebox{0.06\linewidth}{Block-diag.}
        & \includegraphics[width=0.15\linewidth]{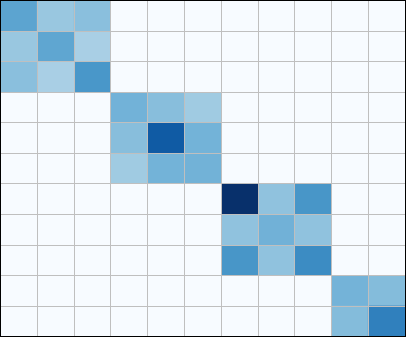}
        & \includegraphics[width=0.15\linewidth]{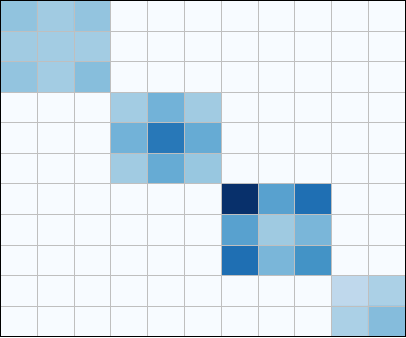}
        & \includegraphics[width=0.15\linewidth]{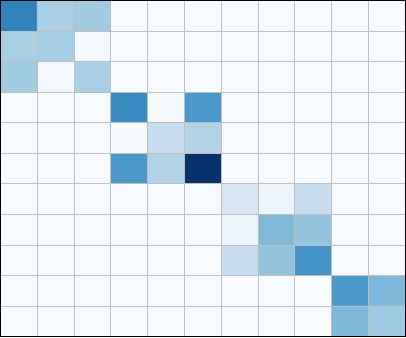}
        \\
        \raisebox{0.06\linewidth}{Tril-Toepl.}
        & \includegraphics[width=0.15\linewidth]{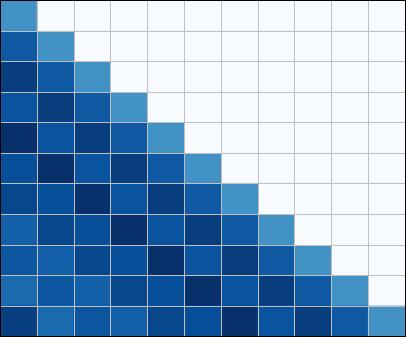}
        & \includegraphics[width=0.15\linewidth]{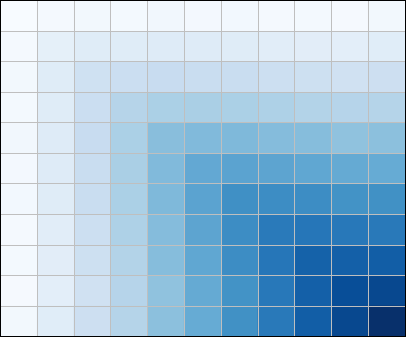}
        & \includegraphics[width=0.15\linewidth]{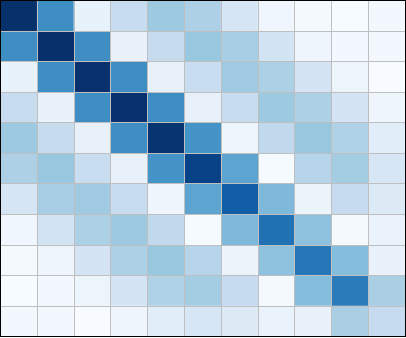}
        \\
        \raisebox{0.06\linewidth}{Triu-Toepl.}
        & \includegraphics[width=0.15\linewidth]{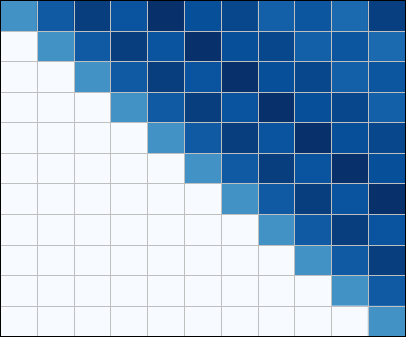}
        & \includegraphics[width=0.15\linewidth]{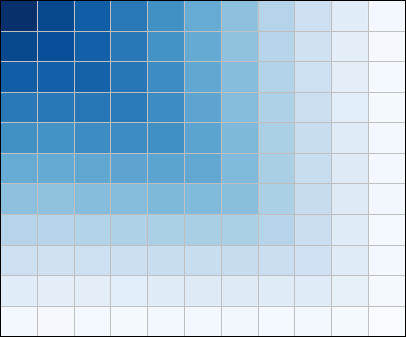}
        & \includegraphics[width=0.15\linewidth]{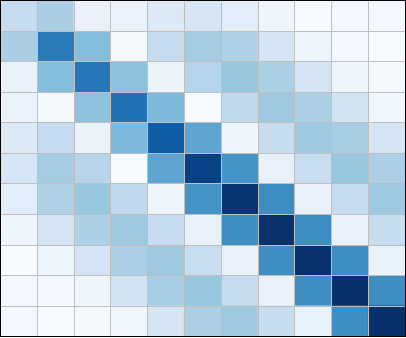}
        \\
        \raisebox{0.06\linewidth}{Hierarchical}
        & \includegraphics[width=0.15\linewidth]{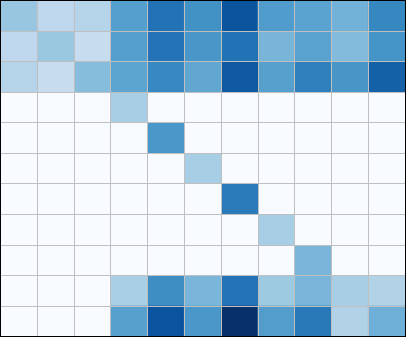}
        & \includegraphics[width=0.15\linewidth]{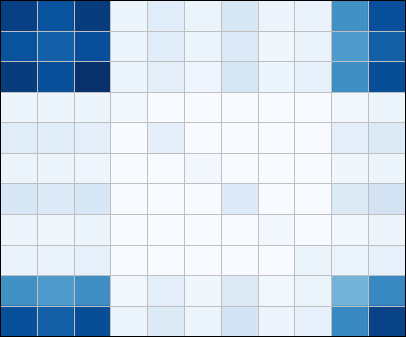}
        & \includegraphics[width=0.15\linewidth]{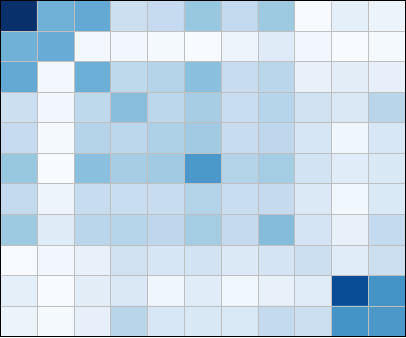}
        \\
        \raisebox{0.06\linewidth}{Sparse Triu.}
        & \includegraphics[width=0.15\linewidth]{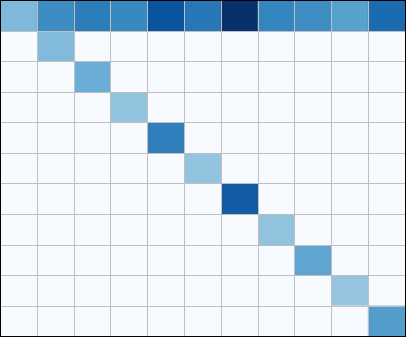}
        & \includegraphics[width=0.15\linewidth]{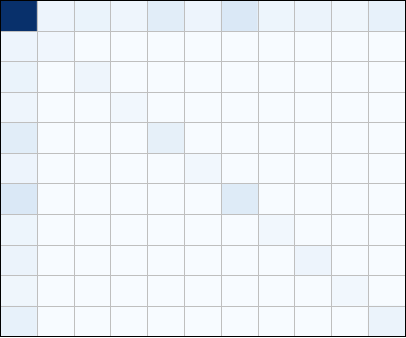}
        & \includegraphics[width=0.15\linewidth]{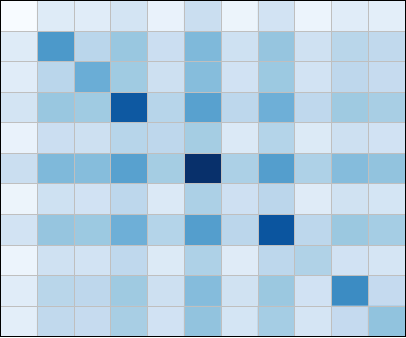}
        \\
        \raisebox{0.06\linewidth}{Sparse Triu.}
        & \includegraphics[width=0.15\linewidth]{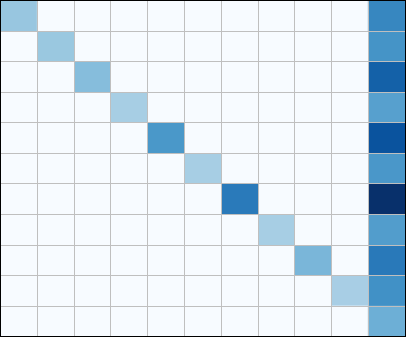}
        & \includegraphics[width=0.15\linewidth]{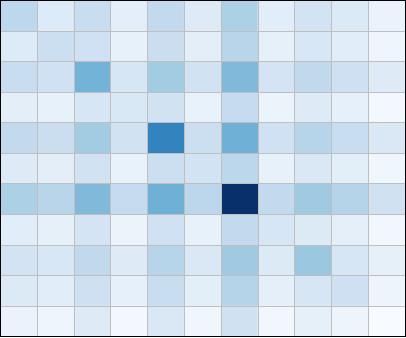}
        & \includegraphics[width=0.15\linewidth]{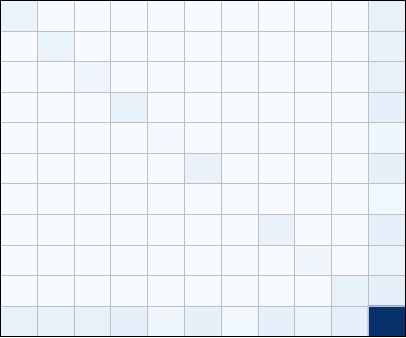}
        \\
        \raisebox{0.06\linewidth}{Sparse Tril.}
        & \includegraphics[width=0.15\linewidth]{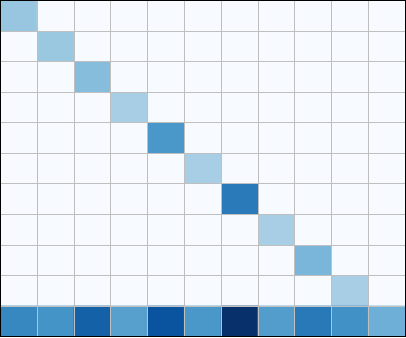}
        & \includegraphics[width=0.15\linewidth]{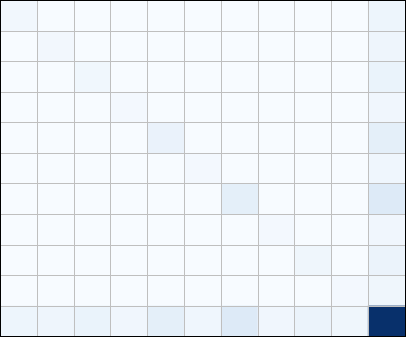}
        & \includegraphics[width=0.15\linewidth]{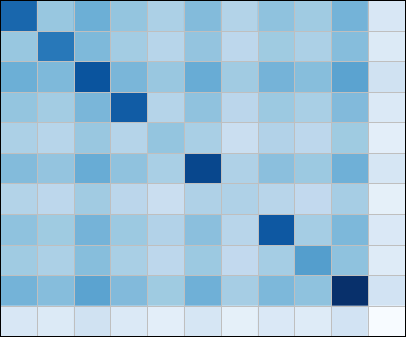}
        \\
        \raisebox{0.06\linewidth}{Sparse Tril.}
        & \includegraphics[width=0.15\linewidth]{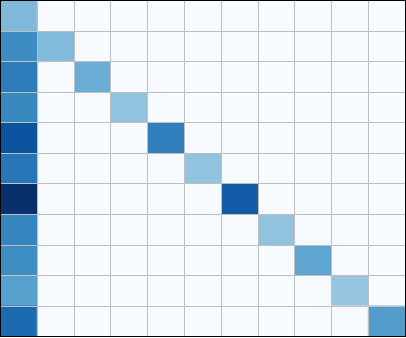}
        & \includegraphics[width=0.15\linewidth]{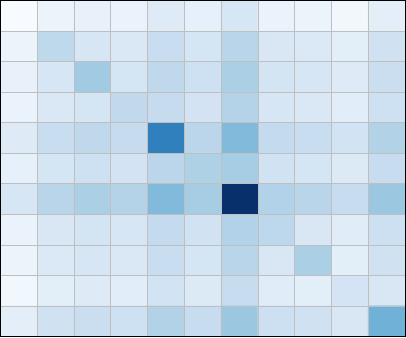}
        & \includegraphics[width=0.15\linewidth]{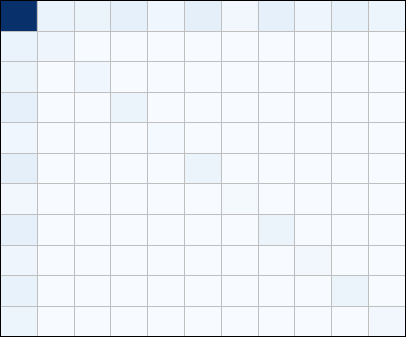}
      \end{tabular}
    \end{small}
    \caption{Illustration of structured matrices (Kronecker factors) supported by SINGD, their self-outer product (approximate inverse Hessian factor), and its inverse (approximate Hessian factor).
      With rank-one triangular matrices $\vK$, we can easily impose a low-rank structure on $\vK\vK^\top$ or $(\vK\vK^{\top})^{-1}$; the latter is difficult to achieve with other approaches.
    }
    \label{fig:structured-matrices}
  \end{minipage}
\end{figure*}

We first propose a new inverse-free update to mimic the behavior of the KFAC update; we call this update IKFAC.
We then show that IKFAC corresponds to a specific setting of INGD.
This bridges the gap between INGD and KFAC and sheds light on the difference between both methods.

Inspired by INGD, we replace matrix inversion with matrix subtraction in a matrix logarithm space, then go back to the original space without explicitly inverting any matrix using a truncated matrix exponential map.
The IKFAC update is related to the KFAC update as we will use
$\vK \smash{\vK^\top}$  and $\vC \smash{\vC^\top} $ to approximate the inverse Kronecker  factors $ \smash{\big( {\vS}_K + \lambda \vI \big)^{-1}}$ and $\smash{\big( {\vS}_C + \lambda \vI \big)^{-1}}$ in KFAC, respectively.
We propose the following IKFAC update with learning rate $\stepsize_1$
for $\vK$ and $\vC$ using a truncated matrix exponential
\begin{align}\label{eq:ikfac_opt}
  \begin{split}
    \vK^{\text{new}}
    &\leftarrow
      \vK  \left( \vI - \nicefrac{\stepsize_1}{2}  {\vm_K} \right)\,,
    \\
    \vC^{\text{new}}
    &\leftarrow
      \vC  \left( \vI - \nicefrac{\stepsize_1}{2}  {\vm_C} \right)\,,
  \end{split}
\end{align}
where $\vH_K \coloneq \smash{\vK^\top} \vU \vK$, $\vH_C \coloneq \smash{\vC^\top} \vG \vC $, ${\vm_K} \coloneq \vH_K + \lambda \smash{\vK^\top}\vK - \vI$, ${\vm_C} \coloneq \vH_C + \lambda \smash{\vC^\top}\vC - \vI$.
This update is inverse- and matrix-decomposition-free.
Since we truncate the matrix exponential $\expm( - \nicefrac{\stepsize_1}{2} \vm_K) \approx ( \vI - \nicefrac{\stepsize_1}{2} \, {\vm_K} )$, $\vm_K$ indeed stays in a matrix logarithm space (see \Cref{app:first_error_analysis}).
The logarithm space allows to impose structural constraints on $\vK$ we discuss in \Cref{sec:structures}.

The following theorem---proof in \Cref{app:proof_of_thm}---formally shows that $\vK \vK^\top$ used in IKFAC is an approximation of $\smash{( \vS _K + \lambda \vI )^{-1}}$ in KFAC at every step even with a truncated matrix exponential.
Similarly, $\vC \smash{\vC^\top}$ is an approximation of $\smash{( \vS _C + \lambda \vI )^{-1}}$.
Thus, IKFAC effectively recovers KFAC up to a first-order accuracy.

\begin{thm}
  \label{thm:kfac_K_part}
  If  $\vK$ is updated according to the IKFAC scheme (\Cref{fig:matDL_opt2}) with the truncation of the matrix exponential and
  these two updates use the same initialization and the same sequence of curvature matrices $\vU$,
  then the product $\vK \smash{\vK^\top}$ has a first-order accuracy of the KFAC update of $\smash{\big(\vS_K  + \lambda \vI \big)^{-1}}$  at each iteration, i.e., \({\vK}   \smash{{\vK}^\top}   =    \smash{\big( {\vS}_K +\lambda \vI \big)^{-1}}  + O(\stepsize_1^2)\).
\end{thm}
\Cref{thm:kfac_K_part} trivially extends to diagonal and block-diagonal structures. I.e., KFAC with diagonal or block-diagonal Kronecker factors is equivalent to IKFAC with diagonal or block-diagonal structure up to first order in $\beta_1 $.

Now, we show that IKFAC is a specific case of INGD,
whose update of $\vK$ without Riemannian momentum ($\alpha_1=0$) is
\begin{align}
  \scalebox{0.75}{%
  $
  \vK^{\text{new}} \!\!\leftarrow    \vK  \left[ \vI_{d_i} - \frac{\stepsize_1}{2 d_o} \big( { \color{red} \mathrm{Tr}(\vH_C)} \vH_K + \lambda {\color{red} \mathrm{Tr}(\vC^\top\vC)} \vK^\top\vK - d_o \vI_{d_i}   \big) \right]
  $
  }
\end{align}
Since $\mathrm{Tr}(\vI_{d_o})=d_o$, $\vH_C \in \real^{d_o \times d_o}$, $\vC \in \real^{d_o \times d_o}$, and $\vK \in \real^{d_i \times d_i}$, we can obtain IKFAC from INGD by simply replacing $\mathrm{Tr}(\vH_C)$ and $\mathrm{Tr}(\smash{\vC^\top}\vC)$ with $\mathrm{Tr}(\vI_{d_o})$:
\begin{align}
  \scalebox{0.75}{%
  $
  \vK^{\text{new}} \leftarrow    \vK  \left[ \vI_{d_i} - \frac{\stepsize_1}{2 d_o} \big( { \color{red} \mathrm{Tr}(\vI_{d_o})} \vH_K + \lambda {\color{red} \mathrm{Tr}(\vI_{d_o})} \vK^\top\vK - d_o \vI_{d_i}    \big) \right] .
  $
  }
\end{align}
This sheds light on the difference between both methods.
In IKFAC (see \Cref{app:first_error_analysis} for details), $\vH_K$ and $\lambda \smash{\vK^\top}\vK$ are used for incorporating KFAC's curvature $\vU$ and damping $\lambda \vI$, respectively.
In contrast, the curvature and damping are \emph{adaptively} incorporated in INGD using $( \mathrm{Tr}(\vH_C)/d_o) \vH_K$ and $(\lambda \mathrm{Tr}(\vC^\top \vC)/d_o) \vK^\top\vK$.
The updates of $\vK$ and $\vC$ are \emph{correlated} in INGD due to the trace terms,
while $\vK$ and $\vC$ are updated independently in IKFAC---just like $\vS_K$ and $\vS_C$ in KFAC.
These trace terms are needed to satisfy the orthonormalization condition of the Fisher matrix \citep{lin2023simplifying}.
They make INGD and SINGD scale-invariant to the Kronecker approximation (see \Cref{app:invariance}) as the approximation is not unique.
In contrast, KFAC and IKFAC are not scale-invariant.
The trace terms together with Riemannian momentum ($\alpha_1>0$) are missing in KFAC and IKFAC.
Our experiments show that they can contribute to stability.

\subsection{Sparse Kronecker Factors for Reducing Memory}
\label{sec:structures}

\begin{figure*}[!t]
  \centering
  \includegraphics[width=\linewidth]{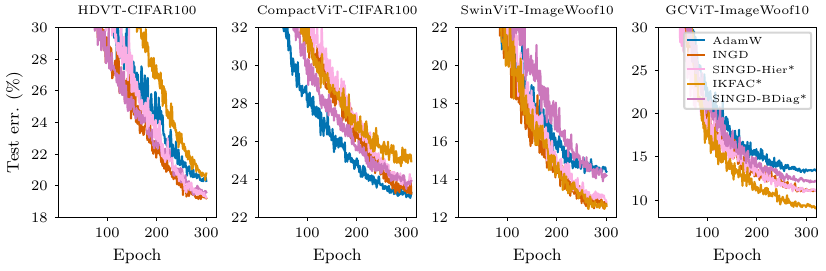}
  \caption{ Test error curves for mixed-precision training in the transformer-based models with BFP-16 on datasets `CIFAR-100' and `ImageWoof-10'.
    SINGD performs as well as INGD while being memory efficient and, including IKFAC and INGD as special cases, outperforms AdamW in most of the cases.
    We omit KFAC since it performs unstably in BFP-16.
    The hierarchical structure often performs as well as the dense structure and outperforms the block-diagonal structure.
  }
  \label{fig:exp}
\end{figure*}

Now, we extend INGD to reduce its memory and iteration cost.
Existing sparse KFAC methods use (block-)diagonal structures for $\vS_K$ and $\vS_C$ \citep{zhang2019which,grosse2023studying}.
In contrast, we propose using sparse Kronecker factors $\vK$ and $\vC$ in INGD and exploiting Lie-algebraic properties in the logarithm space and algebraic sparsity of the Kronecker factors.
This enables more flexible structures (\Cref{fig:structured-matrices}) that potentially achieve better downstream performance than (block-)diagonal structures in $\vS_K$, $\vS_C$.

Other related works are Lie group preconditioners \citep{li2018preconditioner,li2022black} originally derived from directly approximating the Hessian inverse.
Some Hessian-vector-product-based versions of these methods can be expensive and unavailable in pure low-precision settings due to sampling random weights and solving linear systems that are unstable in low precision.
Our approach is \emph{sampling-free} and \emph{available} in pure half-precision settings.

We want to construct sparse factors $\vK$ and $\vC$ without requiring the Kronecker/Hessian approximation ($\vU \otimes \vG $) to be further sparse or structured.
Imposing sparsity often leads to a complicated FIM which makes it difficult to perform NGD due to the FIM inversion.
It is essential to update $\vm_K$ as the logarithm space of $\vK$ to impose sparsity on $\vK$ as the FIM in this (moving) coordinate $\vm_K$ is simplified and becomes an identity matrix due to the orthonormalization condition.
This condition \citep{lin2023simplifying} makes it easy for us to impose a range of sparse structures on $\vK$ through \emph{a unified and inverse-free update rule} (\Cref{fig:matDL_opt}) since we can avoid inverting the Fisher block regarding the sparse structures.
We also exploit the algebraic sparsity in these structures to make our rule more efficient than INGD (\Cref{table:cost}).

We exploit Lie-algebraic properties in the log space of $\vm_K$ to construct sparse structures of $\vK$.
As a general design principle, we consider structures of $\vK$ preserved under (i) elementwise matrix operations (subtraction and scalar multiplication) and (ii) matrix multiplication, which are needed for our updates.
Concretely, we construct a new local reparameterization for $\vK$ at iteration $t$ via
\begin{align*}
  \vK \coloneq \vpsi(\vK_t,\vm_K)
  \coloneq\vK_t \expm \left( \frac{1}{ \sqrt{2d_i}} \, \smash{\hat{\Pi}}_K( \vm_K ) \right)\,,
\end{align*}
where $\smash{\hat{\Pi}}_{K}(\vm_K)$ projects the dense $\vm_K$ to a subspace (identically for $\vC$, but potentially using a different structure $\hat{\Pi}_C$.

Many popular structures such as tri-diagonal matrices do not satisfy our requirements as they are not closed under matrix multiplication.
Moreover, it can be difficult to construct the projection map to satisfy the orthonormalization condition.
One subspace structure satisfying the requirements are upper/lower triangular matrices.
The subspace projection $\smash{\hat{\Pi}_{K}}$ is a weighted extraction map since projecting the logarithm space onto a subspace is like projecting a dense square matrix onto a triangular matrix.
Technically, we use
\begin{align*}
  \vA \coloneq   \vK_t \expm\left( \frac{ \hat{\Pi}_K( \vm_K ) }{\sqrt{2 d_i}}  \right) \otimes \vC_t
\end{align*}
to update $\vK$ at iteration $t$, treating $\vC_t$ and $\vK_t$ as constants.
Given a subspace $\Omega_K \subset \real^{d_i \times d_i}$ in the matrix logarithm space, the subspace projection map $\hat{\Pi}_K: \mathrm{Sym}^{d_i \times d_i} \mapsto \Omega_K $ is specified by satisfying the local orthonormalization condition   of  the Fisher block regarding $\vm_K$:
\begin{align*}
  \fim |_{m_K = \mathbf{0}}
  \coloneq - \mathbb{E}_{w \sim q}\left[  \nabla_{m_K}^2 \log q(\vw \mid \vmu,\vS) \right] \big|_{m_K = \mathbf{0}} = \vI\,,
\end{align*}
with the variational Gaussian $q(\vw\mid\vmu,\vS)$ with mean $\vmu$, precision $\vS\coloneq\vA^{-\top} \vA^{-1}$ and $\mathrm{Sym}^{d_i \times d_i}$ the set of symmetric square real matrices. Similarly, we can obtain $\hat{\Pi}_C$ for $\vC$.

We consider several sparsities and block extensions of triangular matrices illustrated in \Cref{fig:structured-matrices}.
E.g., the subspace projection map for a diagonal structure simply extracts diagonal entries of its input.
As a non-trivial example, the subspace projection map for a lower-triangular structure extracts lower-triangular entries of its input and multiplies the entries below the main diagonal by 2.
\Cref{tab:subspace_proj} summarizes structures and their projection maps mathematically.

Using such a subspace and its projection map, we obtain a structured INGD update (\Cref{fig:matDL_opt}), and similar for IKFAC.
Our approach allows to use more expressive structures than the block-diagonal structure shown in \Cref{fig:structured-matrices}, e.g.\,low-rank, flexible hierarchical, and Toeplitz structures.
While existing methods mainly support low-rank structures.
For an efficient implementation, we only compute and store non-zero entries of $\smash{\hat{\Pi}_{K}}(\vm_K)$ and $\vK$ without explicitly forming dense matrices.
These structures lower not only memory consumption (\Cref{table:storage}), but also the iteration cost (\Cref{table:cost}).

\section{Experiments}
\label{sec:experiment}

We evaluate SINGD on convolutional, transformer, and graph NNs, using mixed-precision training in BFP-16 with KFAC-reduce \citep{eschenhagen2023kronecker} and numerical tricks \citep{dangel2023convolutions} to further reduce memory consumption and iteration cost for convolutions.
The performance metric is test error.
To be memory-efficient, we consider SINGD  with sparse structures such as `diagonal', `block-diagonal', and `hierarchical'.
We also consider IKFAC, INGD (recall SINGD with dense structure becomes INGD), and AdamW as baselines.
All methods except KFAC directly support training in BFP-16.
For KFAC, we have to transform a matrix into FP-32 and then transform its inverse into BFP-16. We find that KFAC performs unstably in BFP-16.
For `VGG' and `ConvMixer', we also consider SGD as a strong baseline, %
We fix momentum to 0.9 and tune other hyper-parameters of each optimizer using random search.
For `VGG' and `ConvMixer', we decrease the learning rate $\stepsize_2$ every 40 epochs.
For `GNN', we use a constant learning rate;  all other models use a cosine learning rate schedule.
We consider KFAC as a strong baseline for the GNN as suggested by \citet{izadi2020optimization}. %
We train the GNN in FP-32 so that KFAC performs stably.
The search space for the random search can be found in \Cref{table:hyperparameters_opt} in \Cref{app:exp}.

\begin{figure*}[t]
  \centering
  \includegraphics[width=\linewidth]{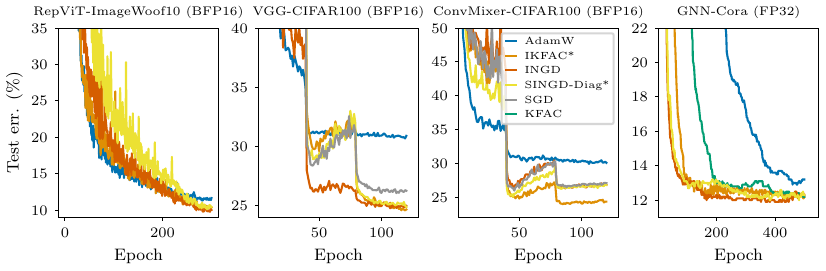}
  \vspace{-0.5cm}
  \caption{
        Test error curves for mixed-precision training in CNN and GNN models on datasets  `ImageWoof-10',  `CIFAR-100'  and `Cora'.
        `Rep-ViT' is a CNN model inspired by transformers.
    SINGD performs as well as INGD while being memory efficient. SINGD including IKFAC and INGD as special cases,  outperforms AdamW on all the models.  The diagonal structure can perform as well as the dense structure on these models.
    KFAC only appears in the rightmost plot since it performs unstably in the other plots due to numerical issues in half-precision settings.
  }
  \label{fig:exp_more}
\end{figure*}

From \Cref{fig:exp} and \ref{fig:exp_more}, we can observe that SINGD, including IKFAC and INGD as special cases, outperforms AdamW in many cases. SINGD  works well for mixed-precision training. We do not show KFAC in the plots as it performs unstably due to numerical issues.
We also observe that  the hierarchical structure often performs as well as the dense structure (INGD) on all the models. In several cases,  the hierarchical structure outperforms the block-diagonal and diagonal structures.
However, on the models shown in \Cref{fig:exp_more}, even the diagonal structure can perform as well as the dense one.
Thus, we can reduce INGD's memory consumption and make SINGD as competitive as AdamW. %
We also train a ViT model on  ``ImageNet-100" to demonstrate the superior performance of SINGD over AdamW in large-scale settings (see ~\Cref{fig:exp_large} in \Cref{app:exp}).

\section{Conclusion}
\label{sec:conclusion}

We propose an inverse-free, memory-efficient natural gradient descent method---SINGD---which addresses the numerical instability and memory inefficiency of  second-order methods like KFAC~\citep{martens2015optimizing}.
The algorithm is an extension of the inverse-free natural gradient (INGD) method from \citet{lin2023simplifying}, whose update relies only on matrix multiplications.
We theoretically establish the algorithm's relation to KFAC by showing that a modification of INGD effectively performs KFAC-like updates and further improve its memory efficiency through sparse Kronecker factors.
We showed that SINGD supports low-precision training and often outperforms AdamW on transformer-based models.
Our work expands the scope of second-order methods to training transformer-based NNs and in low precision, making them more widely applicable.

\section*{Acknowledgements}
Resources used in preparing this research were provided, in part, by the Province of Ontario, the Government of Canada through CIFAR, and companies sponsoring Vector Institute.
Runa Eschenhagen is supported by ARM and the Cambridge Trust.
Richard E. Turner is supported by Google, Amazon, ARM, Improbable and EPSRC grant EP/T005386/1.

\section*{Impact Statement}
This paper presents work whose goal is to advance the field of Machine Learning.
There are many potential societal consequences of our work, none which we feel must be specifically highlighted here.

\bibliography{refs}

\begin{thebibliography}{57}
\providecommand{\natexlab}[1]{#1}
\providecommand{\url}[1]{\texttt{#1}}
\expandafter\ifx\csname urlstyle\endcsname\relax
  \providecommand{\doi}[1]{doi: #1}\else
  \providecommand{\doi}{doi: \begingroup \urlstyle{rm}\Url}\fi

\bibitem[Amari(1998)]{amari1998natural}
Amari, S.-I.
\newblock Natural gradient works efficiently in learning.
\newblock \emph{Neural computation}, 10\penalty0 (2):\penalty0 251--276, 1998.

\bibitem[Bae et~al.(2022)Bae, Ng, Lo, Ghassemi, and Grosse]{bae2022if}
Bae, J., Ng, N., Lo, A., Ghassemi, M., and Grosse, R.~B.
\newblock If influence functions are the answer, then what is the question?
\newblock In \emph{NeurIPS}, 2022.

\bibitem[Botev et~al.(2017)Botev, Ritter, and Barber]{botev2017practical}
Botev, A., Ritter, H., and Barber, D.
\newblock Practical {G}auss-{N}ewton optimisation for deep learning.
\newblock In \emph{ICML}, 2017.

\bibitem[Bradbury et~al.(2018)Bradbury, Frostig, Hawkins, Johnson, Leary, Maclaurin, Necula, Paszke, Vander{P}las, Wanderman-{M}ilne, and Zhang]{jax2018github}
Bradbury, J., Frostig, R., Hawkins, P., Johnson, M.~J., Leary, C., Maclaurin, D., Necula, G., Paszke, A., Vander{P}las, J., Wanderman-{M}ilne, S., and Zhang, Q.
\newblock {JAX}: composable transformations of {P}ython+{N}um{P}y programs, 2018.
\newblock URL \url{http://github.com/google/jax}.

\bibitem[Brown et~al.(2020)Brown, Mann, Ryder, Subbiah, Kaplan, Dhariwal, Neelakantan, Shyam, Sastry, Askell, et~al.]{brown2020gpt3}
Brown, T., Mann, B., Ryder, N., Subbiah, M., Kaplan, J.~D., Dhariwal, P., Neelakantan, A., Shyam, P., Sastry, G., Askell, A., et~al.
\newblock Language models are few-shot learners.
\newblock In \emph{NeurIPS}, 2020.

\bibitem[Dangel(2023)]{dangel2023convolutions}
Dangel, F.
\newblock Convolutions through the lens of tensor networks.
\newblock ar{X}iv 2307.02275, 2023.

\bibitem[Daxberger et~al.(2021)Daxberger, Kristiadi, Immer, Eschenhagen, Bauer, and Hennig]{daxberger2021laplace}
Daxberger, E., Kristiadi, A., Immer, A., Eschenhagen, R., Bauer, M., and Hennig, P.
\newblock {L}aplace redux---effortless {B}ayesian deep learning.
\newblock In \emph{NeurIPS}, 2021.

\bibitem[Dehghani et~al.(2023)Dehghani, Djolonga, Mustafa, Padlewski, Heek, Gilmer, Steiner, Caron, Geirhos, Alabdulmohsin, et~al.]{dehghani2023scaling}
Dehghani, M., Djolonga, J., Mustafa, B., Padlewski, P., Heek, J., Gilmer, J., Steiner, A.~P., Caron, M., Geirhos, R., Alabdulmohsin, I., et~al.
\newblock Scaling vision transformers to 22 billion parameters.
\newblock In \emph{ICML}, 2023.

\bibitem[Eschenhagen et~al.(2023)Eschenhagen, Immer, Turner, Schneider, and Hennig]{eschenhagen2023kronecker}
Eschenhagen, R., Immer, A., Turner, R.~E., Schneider, F., and Hennig, P.
\newblock {K}ronecker-{F}actored {A}pproximate {C}urvature for modern neural network architectures.
\newblock In \emph{{NeurIPS}}, 2023.

\bibitem[George et~al.(2018)George, Laurent, Bouthillier, Ballas, and Vincent]{george2018fast}
George, T., Laurent, C., Bouthillier, X., Ballas, N., and Vincent, P.
\newblock Fast approximate natural gradient descent in a kronecker factored eigenbasis.
\newblock In \emph{NeurIPS}, 2018.

\bibitem[Graves(2011)]{graves2011practical}
Graves, A.
\newblock {Practical variational inference for neural networks}.
\newblock In \emph{NeurIPS}, 2011.

\bibitem[Grosse \& Martens(2016)Grosse and Martens]{grosse2016kronecker}
Grosse, R. and Martens, J.
\newblock A {k}ronecker-factored approximate {f}isher matrix for convolution layers.
\newblock In \emph{{ICML}}, 2016.

\bibitem[Grosse et~al.(2023)Grosse, Bae, Anil, Elhage, Tamkin, Tajdini, Steiner, Li, Durmus, Perez, et~al.]{grosse2023studying}
Grosse, R., Bae, J., Anil, C., Elhage, N., Tamkin, A., Tajdini, A., Steiner, B., Li, D., Durmus, E., Perez, E., et~al.
\newblock Studying large language model generalization with influence functions.
\newblock \emph{arXiv preprint arXiv:2308.03296}, 2023.

\bibitem[Hassani et~al.(2021)Hassani, Walton, Shah, Abuduweili, Li, and Shi]{hassani2021escaping}
Hassani, A., Walton, S., Shah, N., Abuduweili, A., Li, J., and Shi, H.
\newblock Escaping the big data paradigm with compact transformers.
\newblock \emph{arXiv preprint arXiv:2104.05704}, 2021.

\bibitem[Hatamizadeh et~al.(2023)Hatamizadeh, Yin, Heinrich, Kautz, and Molchanov]{hatamizadeh2023global}
Hatamizadeh, A., Yin, H., Heinrich, G., Kautz, J., and Molchanov, P.
\newblock Global context vision transformers.
\newblock In \emph{International Conference on Machine Learning}, pp.\  12633--12646. PMLR, 2023.

\bibitem[Heskes(2000)]{heskes2000natural}
Heskes, T.
\newblock On ``natural'' learning and pruning in multilayered perceptrons.
\newblock \emph{Neural Computation}, 12\penalty0 (4), 2000.

\bibitem[Immer et~al.(2021)Immer, Bauer, Fortuin, R{\"a}tsch, and Emtiyaz]{immer2021scalable}
Immer, A., Bauer, M., Fortuin, V., R{\"a}tsch, G., and Emtiyaz, K.~M.
\newblock Scalable marginal likelihood estimation for model selection in deep learning.
\newblock In \emph{ICML}, 2021.

\bibitem[Izadi et~al.(2020)Izadi, Fang, Stevenson, and Lin]{izadi2020optimization}
Izadi, M.~R., Fang, Y., Stevenson, R., and Lin, L.
\newblock Optimization of graph neural networks with natural gradient descent.
\newblock In \emph{2020 IEEE international conference on big data (big data)}, pp.\  171--179. IEEE, 2020.

\bibitem[Khan \& Lin(2017)Khan and Lin]{khan2017conjugate}
Khan, M. and Lin, W.
\newblock Conjugate-computation variational inference: Converting variational inference in non-conjugate models to inferences in conjugate models.
\newblock In \emph{Artificial Intelligence and Statistics}, pp.\  878--887, 2017.

\bibitem[Khan \& Nielsen(2018)Khan and Nielsen]{khan2018fast}
Khan, M.~E. and Nielsen, D.
\newblock {Fast yet Simple Natural-Gradient Descent for Variational Inference in Complex Models}.
\newblock \emph{arXiv preprint arXiv:1807.04489}, 2018.

\bibitem[Khan \& Rue(2021)Khan and Rue]{khan2021bayesian}
Khan, M.~E. and Rue, H.
\newblock The bayesian learning rule.
\newblock \emph{arXiv preprint arXiv:2107.04562}, 2021.

\bibitem[Khan et~al.(2018)Khan, Nielsen, Tangkaratt, Lin, Gal, and Srivastava]{khan18a}
Khan, M.~E., Nielsen, D., Tangkaratt, V., Lin, W., Gal, Y., and Srivastava, A.
\newblock {Fast and scalable {B}ayesian deep learning by weight-perturbation in {A}dam}.
\newblock In \emph{ICML}, 2018.

\bibitem[Kingma \& Ba(2015)Kingma and Ba]{kingma2014adam}
Kingma, D.~P. and Ba, J.
\newblock Adam: A method for stochastic optimization.
\newblock In \emph{International Conference on Learning Representations}, 2015.

\bibitem[Kipf \& Welling(2016)Kipf and Welling]{kipf2016semi}
Kipf, T.~N. and Welling, M.
\newblock Semi-supervised classification with graph convolutional networks.
\newblock \emph{arXiv preprint arXiv:1609.02907}, 2016.

\bibitem[Kunstner et~al.(2019)Kunstner, Balles, and Hennig]{kunstner2019limitations}
Kunstner, F., Balles, L., and Hennig, P.
\newblock Limitations of the empirical {F}isher approximation for natural gradient descent.
\newblock In \emph{NeurIPS}, 2019.

\bibitem[Li(2022)]{li2022black}
Li, X.
\newblock Black box lie group preconditioners for sgd.
\newblock \emph{arXiv preprint arXiv:2211.04422}, 2022.

\bibitem[Li(2018)]{li2018preconditioner}
Li, X.-L.
\newblock Preconditioner on matrix lie group for sgd.
\newblock In \emph{International Conference on Learning Representations}, 2018.

\bibitem[Lin et~al.(2020)Lin, Schmidt, and Khan]{lin2020handling}
Lin, W., Schmidt, M., and Khan, M.~E.
\newblock Handling the positive-definite constraint in the bayesian learning rule.
\newblock In \emph{ICML}, 2020.

\bibitem[Lin et~al.(2021)Lin, Nielsen, Emtiyaz, and Schmidt]{lin2021tractable}
Lin, W., Nielsen, F., Emtiyaz, K.~M., and Schmidt, M.
\newblock Tractable structured natural-gradient descent using local parameterizations.
\newblock In \emph{ICML}, 2021.

\bibitem[Lin et~al.(2023)Lin, Duruisseaux, Leok, Nielsen, Khan, and Schmidt]{lin2023simplifying}
Lin, W., Duruisseaux, V., Leok, M., Nielsen, F., Khan, M.~E., and Schmidt, M.
\newblock Simplifying momentum-based positive-definite submanifold optimization with applications to deep learning.
\newblock In \emph{ICML}, 2023.

\bibitem[Liu et~al.(2021)Liu, Lin, Cao, Hu, Wei, Zhang, Lin, and Guo]{liu2021swin}
Liu, Z., Lin, Y., Cao, Y., Hu, H., Wei, Y., Zhang, Z., Lin, S., and Guo, B.
\newblock Swin transformer: Hierarchical vision transformer using shifted windows.
\newblock In \emph{Proceedings of the IEEE/CVF international conference on computer vision}, pp.\  10012--10022, 2021.

\bibitem[Loshchilov \& Hutter(2019)Loshchilov and Hutter]{loshchilov2019adamw}
Loshchilov, I. and Hutter, F.
\newblock Decoupled weight decay regularization.
\newblock In \emph{ICLR}, 2019.

\bibitem[Lu et~al.(2022)Lu, Xie, Liu, and Zhang]{lu2022bridging}
Lu, Z., Xie, H., Liu, C., and Zhang, Y.
\newblock Bridging the gap between vision transformers and convolutional neural networks on small datasets.
\newblock \emph{Advances in Neural Information Processing Systems}, 35:\penalty0 14663--14677, 2022.

\bibitem[Martens(2014)]{martens2014new}
Martens, J.
\newblock New insights and perspectives on the natural gradient method.
\newblock \emph{JMLR}, 21\penalty0 (146), 2014.

\bibitem[Martens \& Grosse(2015)Martens and Grosse]{martens2015optimizing}
Martens, J. and Grosse, R.
\newblock {Optimizing neural networks with {K}ronecker-factored approximate curvature}.
\newblock In \emph{ICML}, 2015.

\bibitem[Martens et~al.(2018)Martens, Ba, and Johnson]{martens2018kroneckerfactored}
Martens, J., Ba, J., and Johnson, M.
\newblock {K}ronecker-factored curvature approximations for recurrent neural networks.
\newblock In \emph{{ICLR}}, 2018.

\bibitem[Micikevicius et~al.(2018)Micikevicius, Narang, Alben, Diamos, Elsen, Garcia, Ginsburg, Houston, Kuchaiev, Venkatesh, and Wu]{micikevicius2018mixed}
Micikevicius, P., Narang, S., Alben, J., Diamos, G., Elsen, E., Garcia, D., Ginsburg, B., Houston, M., Kuchaiev, O., Venkatesh, G., and Wu, H.
\newblock Mixed precision training.
\newblock In \emph{International Conference on Learning Representations (ICLR)}, 2018.

\bibitem[Osawa et~al.(2019)Osawa, Swaroop, Khan, Jain, Eschenhagen, Turner, and Yokota]{osawa2019practical}
Osawa, K., Swaroop, S., Khan, M. E.~E., Jain, A., Eschenhagen, R., Turner, R.~E., and Yokota, R.
\newblock {Practical deep learning with Bayesian principles}.
\newblock In \emph{NeurIPS}, 2019.

\bibitem[Osawa et~al.(2023)Osawa, Li, and Hoefler]{osawa2023pipefisher}
Osawa, K., Li, S., and Hoefler, T.
\newblock {P}ipe{F}isher: Efficient training of large language models using pipelining and {F}isher information matrices.
\newblock In \emph{MLSys}, 2023.

\bibitem[Osborne(1992)]{osborne1992fisher}
Osborne, M.~R.
\newblock Fisher's method of scoring.
\newblock \emph{International Statistical Review/Revue Internationale de Statistique}, pp.\  99--117, 1992.

\bibitem[Paszke et~al.(2019)Paszke, Gross, Massa, Lerer, Bradbury, Chanan, Killeen, Lin, Gimelshein, Antiga, et~al.]{paszke2019pytorch}
Paszke, A., Gross, S., Massa, F., Lerer, A., Bradbury, J., Chanan, G., Killeen, T., Lin, Z., Gimelshein, N., Antiga, L., et~al.
\newblock {PyTorch}: An imperative style, high-performance deep learning library.
\newblock In \emph{NeurIPS}, 2019.

\bibitem[Radford et~al.(2019)Radford, Wu, Child, Luan, Amodei, Sutskever, et~al.]{radford2019gpt2}
Radford, A., Wu, J., Child, R., Luan, D., Amodei, D., Sutskever, I., et~al.
\newblock Language models are unsupervised multitask learners.
\newblock \emph{OpenAI blog}, 1\penalty0 (8):\penalty0 9, 2019.

\bibitem[Robbins \& Monro(1951)Robbins and Monro]{robbins1951stochastic}
Robbins, H. and Monro, S.
\newblock {A Stochastic Approximation Method}.
\newblock \emph{The Annals of Mathematical Statistics}, 1951.

\bibitem[Schraudolph(2002)]{schraudolph2002fast}
Schraudolph, N.~N.
\newblock Fast curvature matrix-vector products for second-order gradient descent.
\newblock \emph{Neural computation}, 14\penalty0 (7), 2002.

\bibitem[Simonyan \& Zisserman(2014)Simonyan and Zisserman]{simonyan2014very}
Simonyan, K. and Zisserman, A.
\newblock Very deep convolutional networks for large-scale image recognition.
\newblock \emph{arXiv preprint arXiv:1409.1556}, 2014.

\bibitem[Smyth(1996)]{smyth1996partitioned}
Smyth, G.~K.
\newblock Partitioned algorithms for maximum likelihood and other non-linear estimation.
\newblock \emph{Statistics and Computing}, 6:\penalty0 201--216, 1996.

\bibitem[Smyth(2015)]{smyth2015optimization}
Smyth, G.~K.
\newblock Optimization and nonlinear equations.
\newblock \emph{Statistics reference online}, 1:\penalty0 1--9, 2015.

\bibitem[Tan(2022)]{tan2022analytic}
Tan, L.~S.
\newblock Analytic natural gradient updates for cholesky factor in gaussian variational approximation.
\newblock \emph{arXiv preprint arXiv:2109.00375}, 2022.

\bibitem[Thompson et~al.(2020)Thompson, Greenewald, Lee, and Manso]{thompson2020computational}
Thompson, N.~C., Greenewald, K., Lee, K., and Manso, G.~F.
\newblock The computational limits of deep learning.
\newblock 2020.

\bibitem[Touvron et~al.(2023)Touvron, Lavril, Izacard, Martinet, Lachaux, Lacroix, Rozi{\`e}re, Goyal, Hambro, Azhar, et~al.]{touvron2023llama}
Touvron, H., Lavril, T., Izacard, G., Martinet, X., Lachaux, M.-A., Lacroix, T., Rozi{\`e}re, B., Goyal, N., Hambro, E., Azhar, F., et~al.
\newblock {LLaMA}: Open and efficient foundation language models.
\newblock \emph{arXiv preprint arXiv:2302.13971}, 2023.

\bibitem[Trockman \& Kolter(2023)Trockman and Kolter]{trockman2023patches}
Trockman, A. and Kolter, J.~Z.
\newblock Patches are all you need?
\newblock \emph{Transactions on Machine Learning Research}, 2023.

\bibitem[Vaswani et~al.(2017)Vaswani, Shazeer, Parmar, Uszkoreit, Jones, Gomez, Kaiser, and Polosukhin]{vaswani2017attention}
Vaswani, A., Shazeer, N., Parmar, N., Uszkoreit, J., Jones, L., Gomez, A.~N., Kaiser, {\L}., and Polosukhin, I.
\newblock Attention is all you need.
\newblock In \emph{NIPS}, 2017.

\bibitem[Wang et~al.(2023)Wang, Chen, Lin, Pu, and Ding]{wang2023repvit}
Wang, A., Chen, H., Lin, Z., Pu, H., and Ding, G.
\newblock Repvit: Revisiting mobile cnn from vit perspective.
\newblock \emph{arXiv preprint arXiv:2307.09283}, 2023.

\bibitem[Wang et~al.(2019)Wang, Grosse, Fidler, and Zhang]{wang2019eigendamage}
Wang, C., Grosse, R., Fidler, S., and Zhang, G.
\newblock Eigendamage: Structured pruning in the kronecker-factored eigenbasis.
\newblock In \emph{ICML}, 2019.

\bibitem[Wang(2010)]{wang2010fisher}
Wang, Y.
\newblock {F}isher scoring: An interpolation family and its {M}onte {C}arlo implementations.
\newblock \emph{Comput. Stat. Data Anal.}, 54\penalty0 (7), 2010.

\bibitem[Zhang et~al.(2018)Zhang, Sun, Duvenaud, and Grosse]{zhang2018noisy}
Zhang, G., Sun, S., Duvenaud, D., and Grosse, R.
\newblock Noisy natural gradient as variational inference.
\newblock In \emph{ICML}, 2018.

\bibitem[Zhang et~al.(2019)Zhang, Li, Nado, Martens, Sachdeva, Dahl, Shallue, and Grosse]{zhang2019which}
Zhang, G., Li, L., Nado, Z., Martens, J., Sachdeva, S., Dahl, G.~E., Shallue, C.~J., and Grosse, R.~B.
\newblock Which algorithmic choices matter at which batch sizes? {I}nsights from a noisy quadratic model.
\newblock In \emph{{N}eur{IPS}}, 2019.

\end{thebibliography}
\bibliographystyle{icml2024}

\newpage
\appendix
\onecolumn

\section{
space and time complexity
}
\label{app:cost}

\begin{table*}[ht]
  \begin{minipage}[t]{\linewidth}
  \centering
  \resizebox{1 \textwidth}{!}{   %
   \begin{tabular}{c c c c c c}
      \toprule
      \begin{tabular}{c}
                     \\
      \end{tabular} &
      Method &
      \begin{tabular}{c}
                 $	\triangle \vmu $ (descent direction)   \\

      \end{tabular} &
      \begin{tabular}{c}
                Update $\vS_K$ or $\vK$ \\

      \end{tabular} &
      \begin{tabular}{c}
           Update  $\vS_C$ or $\vC$ \\

      \end{tabular} &
      \begin{tabular}{c}
           $\nabla_\mu  \ell $ (BackProp) \\

      \end{tabular}
      \\
                  \midrule
      \multirow{2}{*}{
        \begin{tabular}{c}
                      \\
                   Iteration Cost
      \end{tabular}}
      & KFAC   & $O(d_i^2 d_o + d_o^2 d_i )$ & $O(\frac{1}{T} (m d_i^2 +  d_i^3) ) $     & $O(\frac{1}{T} (m d_o^2 +  d_o^3) ) $    & $O(m d_i d_o  )$     \\
                  &  INGD/SINGD (Dense)      & $O(d_i^2 d_o + d_o^2 d_i )$   &   $O(\frac{1}{T}  (m d_i^2 + d_i^3)  ) $    &  $O(\frac{1}{T} (m d_o^2 +  d_o^3) ) $  &  $O(m d_i d_o   )$       \\
                  &  SINGD (Block-Diag. with block size $k$)     & $O(kd_i  d_o   )$   &   $O(\frac{1}{T}  (km d_i )  ) $    &  $O(\frac{1}{T} (km d_o ) ) $  &  $O(m d_i d_o   )$       \\
                  &  SINGD (Toeplitz)      & $O(d_i  d_o \log (d_o d_i)   )$   &   $O(\frac{1}{T}  (m d_i \log d_i )  ) $    &  $O(\frac{1}{T} (m d_o \log d_o ) ) $  &  $O(m d_i d_o   )$       \\
                   & SINGD (Rank-1 Triangular)  & $O(d_i  d_o    )$   &   $O(\frac{1}{T}  (m d_i   )  ) $    &  $O(\frac{1}{T} (m d_o   ) ) $  &  $O(m d_i d_o   )$       \\
                    & SINGD (Hierarchical with parameter $k$)  &  $O( k  d_i d_o)$  &   $O( \frac{1}{T}  (k m d_i) )$      &  $O( \frac{1}{T}  (k m d_o) )$ & $O(m d_i d_o   )$     \\

      & AdamW      & $O(d_i d_o)$   &  NA  & NA  &   $O(m d_i d_o   )$  \\
                  \bottomrule \\
    \end{tabular}
   }
  \vspace{-0.5cm}
  \caption{
  Iteration cost for a non-weight-sharing layer, where $m$ is the size of a mini-batch and $\vmu \in \real^{d_i \times d_o}$ is a learnable weight matrix. We assume factors $\vK$ and $\vC$ use the same structure.
    }
    \label{table:cost}
  \end{minipage}
  \end{table*}

\begin{table*}[ht]
  \begin{minipage}[t]{\linewidth}
  \centering
  \resizebox{0.8 \textwidth}{!}{   %
   \begin{tabular}{c c c c c}
      \toprule
      \begin{tabular}{c}
                     \\
      \end{tabular} &
      Method &
      \begin{tabular}{c}
                    $\nabla_\mu \ell \odot \nabla_\mu \ell$ \\

      \end{tabular} &
      \begin{tabular}{c}
                 $\vS_K$ or $\vK$ \\

      \end{tabular} &
      \begin{tabular}{c}
              $\vS_C$ or $\vC$ \\

      \end{tabular}
      \\
                  \midrule
      \multirow{2}{*}{
        \begin{tabular}{c}
                      \\
                   Memory Usage
      \end{tabular}}
      & KFAC   & NA  & $O(d_i^2)$    &  $O(d_o^2)$        \\
                  & INGD/SINGD (Dense)  & NA    & $O(d_i^2)$    & $O(d_o^2)$          \\
            & SINGD (Block-Diag. with block size $k$)  & NA    & $O(kd_i)$    & $O(kd_o)$          \\
            & SINGD (Toeplitz) & NA & $O(d_i)$ & $O(d_o)$          \\
          & SINGD (Rank-1 Triangular) & NA & $O(d_i)$ & $O(d_o)$          \\
         & SINGD (Hierarchical with parameter $k$) & NA & $O(k d_i)$ & $O(k d_o)$          \\
      & AdamW      & $O(d_i d_o)$   &  NA    &   NA \\
                  \bottomrule \\
    \end{tabular}
   }
  \vspace{-0.5cm}
  \caption{
  Additional Storage
    }
  \label{table:storage}
  \end{minipage}
  \end{table*}

\section{Details of the Experiments}
\label{app:exp}

\begin{figure}[H]
  \centering
  \includegraphics[width=\linewidth]{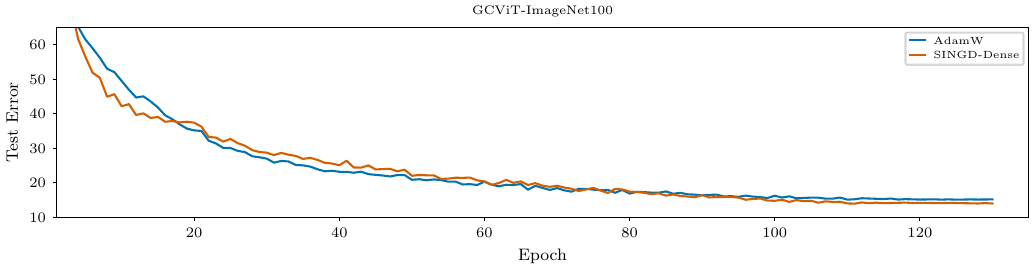}
  \vspace{-0.5cm}
  \caption{
        Test error curves for mixed-precision training on a GCViT model on dataset  `ImageNet-100'.
    SINGD has a similar iteration cost as AdamW while achieving better performance.
  }
  \label{fig:exp_large}
\end{figure}

To demonstrate the robustness and memory efficiency of our method,
we consider image classification tasks with transformer-based models   such as   ``Compact-ViT" \citep{hassani2021escaping}, ``Swin-ViT" \citep{liu2021swin}, ``GC-ViT" \citep{hatamizadeh2023global}, and ``HDVT'' \citep{lu2022bridging}. We also consider  convolution-based models  such as ``VGG'' \citep{simonyan2014very}, ``ConvMixer'' \citep{trockman2023patches},  and ``Rep-ViT" \citep{wang2023repvit}.
We train these models on datasets ``CIFAR-100" and ``ImageWoof-10". Note that ``Rep-ViT" is a CNN model inspired by transformers while  ``Compact-ViT" is a data-efficient
transformer using convolutional tokenization. We also consider a graph convolution model   \citep{kipf2016semi} denoted by
``GNN'' for node classification on dataset ``Cora".
We also train a ViT model on ``ImageNet-100" (\url{ https://www.kaggle.com/datasets/ambityga/imagenet100}) to demonstrate the performance of SINGD in large-scale settings (see Fig.~\ref{fig:exp_large}).

\subsection{Hyper-parameter Tuning}

 \begin{table*}[!htbp]
	\centering
	\footnotesize
	\begin{tabular}{c c c c}
		\toprule
		\begin{tabular}{c}
		Hyperparameter
		\end{tabular} &
		\begin{tabular}{c}
                Meaning
		\end{tabular} &
		\begin{tabular}{c}
		KFAC/IKFAC/SINGD in \Cref{fig:matDL_opt} and \ref{fig:matDL_opt3}
		\end{tabular} &
		\begin{tabular}{c}
		AdamW in \Cref{fig:matDL_opt3}
		\end{tabular} \\
                \midrule
		\multirow{1}{*}{
		  \begin{tabular}{c}
                    $\stepsize_2$
		\end{tabular}}
		& Standard stepsize       &  Tuned &  Tuned \\
                \midrule
		\multirow{1}{*}{
		  \begin{tabular}{c}
                    $\alpha_2$
		\end{tabular}}
		&   Standard momentum weight  & 0.9   & 0.9   \\
                \midrule
		\multirow{1}{*}{
		  \begin{tabular}{c}
                    $\gamma$
		\end{tabular}}
		&  (L2) weight decay       &  Tuned &  Tuned \\
                  \midrule
		\multirow{1}{*}{
		  \begin{tabular}{c}
                    $\lambda$
		\end{tabular}}
		& Damping    &  Tuned &   Tuned \\
                    \midrule
		\multirow{1}{*}{
		  \begin{tabular}{c}
                    $\stepsize_1$
		\end{tabular}}
		&  Stepsize for preconditioner       & Tuned & Tuned \\
                      \midrule
		\multirow{1}{*}{
		  \begin{tabular}{c}
                    $\alpha_1$
		\end{tabular}}
		& Riemannian Momentum     & (SINGD only) Tuned & NA \\
                \bottomrule \\
	\end{tabular}
   \caption{Hyperparameters  used for a random search.
   }
\label{table:hyperparameters_opt}
      \end{table*}

\begin{table}[!h]
  \centering
  \caption{Peak memory and run time of different optimizers for GCViT on ImageWoof10 (\Cref{fig:exp}, \emph{right}).
    Parenthesized values are normalized relative to SGD.
    For this vision transformer task, we observe that the backpropagation dominates both run time and memory.
    In this setting, all our methods as well as INGD have basically no run time and memory overhead compared to the first-order methods.
    INGD and our proposed methods are even able to beat AdamW and SGD in terms of test error.
    INGD, KFAC and SINGD update their preconditioner every $T=5$ iterations.
  }
  \label{tab:performance-cgvit-imagewoof}
  \begin{small}
    \begin{tabular}{lcc}
      \toprule
      \multirow{2}{*}{\textbf{Method}} & \textbf{Peak memory} & \textbf{Training time}
      \\
                                       & \textbf{[GiB]} & \textbf{[min]}
      \\
      \midrule
      SGD (BFP-16) & 15.6 (1.00\,x) & 190 (1.00\,x)
      \\
      AdamW (BFP-16) & 15.7  (1.00\,x) & 191 (1.01\,x)
      \\
      SINGD-Diag* (BFP-16) & 15.8 (1.02\,x) & 200 (1.06\,x)
      \\
      IKFAC* (BFP-16) & 16.0 (1.02\,x) & 197 (1.04\,x)
      \\
      INGD (BFP-16) & 16.0 (1.02\,x) & 203 (1.07\,x)
      \\
      KFAC (FP-32) & 16.0 (1.02\,x) & 359 (1.89\,x)
      \\
      \bottomrule
    \end{tabular}
  \end{small}
\end{table}

\begin{figure*}[!h]
\center
	\fbox{
			\begin{minipage}{.47\textwidth}
             \textbf{INGD}
		\begin{algorithmic}[1]
               \STATE
            \footnotesize   Each $T$ iter., update   \scalebox{0.7}{ $\vm_K$, $\vm_C$, $\vK$, $\vC$}   \\
                Obtain  $\vmu_{AA} \otimes \vmu_{GG}$ to approximate $\nabla_\mu^2 \ell(\vmu)$ \\
             \scalebox{0.8}{   $\vm_K  \leftarrow\alpha_1 \vm_K + \frac{1}{2d}(\mathrm{Tr}(\vH_C) \vH_K +  c^2 \vK^T\vK- d\vI_p )$} \\
             \scalebox{0.8}{  $\vm_C  \leftarrow\alpha_1 \vm_C + \frac{1}{2p}(\mathrm{Tr}(\vH_K) \vH_C +  \kappa^2 \vC^T\vC- p\vI_d )$} \\
             \scalebox{0.8}{  $\vK \leftarrow \vK \MExp(-\stepsize_1 \vm_K) \approx \vK (\vI_p-\stepsize_1 \vm_K)$} \\
            \scalebox{0.8}{  $\vC \leftarrow \vC \MExp(-\stepsize_1 \vm_C) \approx \vC (\vI_d-\stepsize_1 \vm_C)$} \\

            \STATE
          \scalebox{0.74}{    $\vM_\mu  \leftarrow \alpha_2 \vM_\mu  + \vC\vC^{T} \mathrm{vec}^{-1}( \nabla_\mu \ell(\vmu)) \vK\vK^{T} + \gamma \mathrm{vec}^{-1} ( \vmu )   $}
             \STATE
            \scalebox{0.8}{   $ \vmu \leftarrow \vmu - \stepsize_2    \mathrm{vec}(\vM_\mu )   $}
				\end{algorithmic}
	\end{minipage}
	}
   \fbox{
			\begin{minipage}{.46\textwidth}
		\textbf{AdamW Optimizer}

				\begin{algorithmic}[1]
               \STATE
            \footnotesize   At iter. $t$, update  \scalebox{0.9}{  $\vm_s$, $\vs$}  \\
            Use \scalebox{0.7}{$\left( \nabla_\mu \ell(\vmu) \right)^2$} to approximate \scalebox{0.7}{$\mathrm{diag}\left(\nabla_\mu^2 \ell(\vmu)\right)$} \\
          \scalebox{0.9}{       $\vm_s \leftarrow (1-\beta_1) \vm_s + \beta_1 \left( \nabla_\mu \ell(\vmu) \right)^2  $} \\
        \scalebox{0.95}{    $ \vs^2 \leftarrow \nicefrac{\vm_s}{ (1- (1-\beta_1)^t) } $ } \\
        \scalebox{0.95}{     $\vs \leftarrow \sqrt{\vs^2} + \lambda $ } \\
        \STATE
        \scalebox{0.8}{    $\vm_\mu  \leftarrow \alpha_2 \vm_\mu  + (1-\alpha_2)  \nabla_\mu \ell(\vmu)    $} \\
          \scalebox{0.8}{    $ \vM_\mu \leftarrow \vs^{-1} \vm_\mu /\big(1-\alpha_2^t\big)     $}
             \STATE
           \scalebox{0.9}{    $ \vmu \leftarrow \vmu - \stepsize_2   \vM_\mu    + \gamma  \vmu $}

				\end{algorithmic}
	\end{minipage}
	}
 \vspace{-0.1cm}   \caption{
 Baseline methods in the same notation
 for a hyperparameter search.
}
\label{fig:matDL_opt3}
\end{figure*}

\section{ Connection between IKFAC and   KFAC  }
\label{app:first_error_analysis}

To relate to the KFAC method,
we now show that $  \vK^{\text{new}} \big(\vK^{\text{new}}\big)^\top$ is an approximation of  $ \big( \vS _K^{\text{new}} + \lambda \vI \big)^{-1}$ at a new step of our  scheme. For simplicity, we first assume
$  \vK  \vK^\top$ exactly equals to $\left( \vS_K^{\text{cur}} + \lambda \vI \right)^{-1}$ at the current step. Later, we will relax this assumption and prove that $ \vK  \vK^\top$ is an approximation of  $\left( \vS_K + \lambda \vI \right)^{-1}$ at every step as stated in Theorem~\ref{thm:kfac_K_part}. %
For notation simplicity, we denote $\bar{\vS}_K\coloneq \vS_K + \lambda \vI$.
The update of $\vS_K$ with damping $\lambda \vI $ can be reexpressed as an update of $\bar{\vS}_K$:
\begin{align*}
 \left( \vS_K^{\text{new}} + \lambda \vI \right) =    \bar{\vS}_K^{\text{new}} \leftarrow  (1-\stepsize_1) \bar{\vS}_K^{\text{cur}} + \stepsize_1 \left( \vU +\lambda \vI \right) .
\end{align*}

Since $\hat{\vS}_K^{\text{cur}} = \vK^{-T}\vK^{-1}$ by our assumption,  we can express update of $\vS_K$ in terms of $\vK$ as follows.
\resizebox{0.95\linewidth}{!}{
  \begin{minipage}[t]{\linewidth}
\begin{align*}
\bar{\vS}_K^{\text{new}} \leftarrow  (1-\stepsize_1) \bar{\vS}_K^{\text{cur}} + \stepsize_1 \left( \vU + \lambda \vI \right) & =    \vK^{-T} \left( \vI + \stepsize_1 \left( \vK^\top \vU   \vK + \lambda \vK^\top\vK - \vI \right)  \right) \vK^{-1} =   \vK^{-T} \left( \vI + \stepsize_1 \vm_K  \right) \vK^{-1}
\end{align*} \end{minipage}
}

$\bar{\vS}_K^{\text{new}}$ in the KFAC update can be approximated as below, where we consider $\vI + \stepsize_1 {\vm_K}$  as an approximate of the matrix exponential $\expm (\stepsize_1 {\vm_K})\approx \vI +\stepsize_1 {\vm_K}$ and notice that ${\vm_K} $ is symmetric.
\resizebox{0.95\linewidth}{!}{
  \begin{minipage}[t]{\linewidth}
 \begin{align*}
    \bar{\vS}_K^{\text{new}} =  \vK^{-T} \left( \vI +  \stepsize_1  {\vm_K} \right)   \vK^{-1} \approx \vK^{-T}   \expm \left(    \stepsize_1   {\vm_K} \right)    \vK^{-1} = \vK^{-T}   \expm \Big(   \frac{\stepsize_1 }{2}   {\vm_K} \Big)^\top  \expm \Big(   \frac{\stepsize_1 }{2}   {\vm_K} \Big)    \vK^{-1}.
\end{align*} \end{minipage}
}

Informally, we can see that $\vK^{\text{new}} \big( \vK^{\text{new}} \big)^\top $ approximates   $\big( \bar{\vS}_K^{\text{new}} \big)^{-1} $ by using  the matrix exponential. We can see that ${\vm_K}$ stays in a matrix logarithm space.

\resizebox{0.95\linewidth}{!}{
  \begin{minipage}[t]{\linewidth}
\begin{align*}
    \left(  \bar{\vS}_K^{\text{new}} \right)^{-1} \approx   \vK \expm \Big(   {\color{red}-} \frac{\stepsize_1}{2}  {\vm_K} \Big) \expm \Big(   {\color{red}-} \frac{\stepsize_1}{2}  {\vm_K} \Big)^\top \vK^\top \approx \vK  \Big( \vI - \frac{\stepsize_1}{2}  {\vm_K} \Big)   \Big( \vI - \frac{\stepsize_1}{2}  {\vm_K} \Big)^{T} \vK^\top =\vK^{\text{new}} \big( \vK^{\text{new}} \big)^\top
\end{align*}
\end{minipage}
}

Theorem~\ref{thm:kfac_K_part} formally shows that $\vK  \vK^\top$ used in our update is an approximation of $\Big( \vS _K + \lambda \vI \Big)^{-1}$ in the   KFAC update for every step even when the truncation of the matrix exponential is employed.

\section{Proof of Theorem~\ref{thm:kfac_K_part} }
\label{app:proof_of_thm}
We first consider the following lemmas in order to prove Theorem~\ref{thm:kfac_K_part}.

Recall that we denote
$\bar{\vS}_K\coloneq \vS_K + \lambda \vI $. For notation simplicity, we will drop the subscript $K$ in this section and use $\bar{\vS}_t$ to denote $\bar{\vS}_K$ at iteration $t$.
Notice that $\bar{\vS}_t$ is non-singular at each iteration $t$ so that we can inverse it in the original KFAC update (see \Cref{fig:matDL_opt2}).

\begin{lemma}
\label{lemma:kfac_identity}
Consider the following update in the original KFAC update at iteration $t$.
\begin{align*}
    \bar{\vS}_t \coloneq(1-\stepsize_1) \bar{\vS}_{t-1} + \stepsize_1 \big( \hat{\vU}_{t-1} +\lambda \vI \big)
\end{align*} where $\vS_t$ is the factor $\vS_K$ used in the original KFAC update, $\stepsize_1$ is known as the weight of the moving average, and $\hat{\vU}_{t-1}$ is a curvature matrix.

The initial factor $ \bar{\vS}_0$ can be decomposed as  $ \bar{\vS}_0 = \hat{\vK}_0^{-T}  \hat{\vK}_0^{-1} $ since $ \bar{\vS}_0$ as a preconditioning factor is symmetric positive definite.

Define $\hat{\vN}_i\coloneq  \hat{\vK}_0^{T} \hat{\vU}_{i}  \hat{\vK}_0 +\lambda \hat{\vK}_0^{T} \hat{\vK}_0 - \vI $.

The Kronecker factor can be reexpressed as
\begin{align*}
     \bar{\vS}_t =  \hat{\vK}_0^{-T} \left( \vI + \stepsize_1 \sum_{i=0}^{t-1} \hat{\vN}_i \right) \hat{\vK}_0^{-1} + O(\stepsize_1^2)
\end{align*}

\end{lemma}

\begin{lemma}
\label{lemma:ours_identity}
Consider the following update in our inverse-free KFAC at iteration $t$.
\begin{align*}
     {\vK}_t \coloneq  {\vK}_{t-1} \left( \vI - \frac{\stepsize_1}{2} \left( {\vK}_{t-1}^\top  {\vU}_{t-1}   {\vK}_{t-1} + \lambda {\vK}_{t-1}^\top {\vK}_{t-1}  - \vI\right) \right)
\end{align*} where ${\vK}_{t-1}^\top  {\vU}_{t-1}  {\vK}_{t-1}$ is used in our update and $ {\vU}_{t-1}$ is a curvature matrix.

Define $ {\vN}_i\coloneq  {\vK}_i^\top  {\vU}_i  {\vK}_i + \lambda {\vK}_i^\top   {\vK}_i - \vI$.

Our update of $ {\vK}$ can be reexpressed as
\begin{align*}
     {\vK}_t =  {\vK}_0 \left( \vI - \frac{\stepsize_1}{2} \sum_{i=0}^{t-1}  {\vN}_i \right) + O(\stepsize_1^2)
\end{align*}

Moreover, the product $  {\vK}  {\vK}^\top  $ can be reexpressed as
\begin{align*}
    {\vK}_t      {\vK}_t^\top =   {\vK}_0 \left( \vI -  \stepsize_1 \sum_{i=0}^{t-1}  {\vN}_i  \right)  {\vK}_0^\top + O(\stepsize_1^2)
\end{align*}

\end{lemma}

Lemma \ref{lemma:kfac_and_ours} is useful to establish a relationship between the KFAC update and our inverse-free update.

\begin{lemma}
\label{lemma:kfac_and_ours}
If we use the same sequence of curvature matrices in both the original KFAC update and our  update such as  $\hat{\vU}_{i}= \vU_{i}$ for each iteration $i$ and $\hat{\vK}_0 = {\vK}_0$ are used on the initialization, we have
the following expression.
\begin{align*}
     {\vN}_i =\hat{\vN}_i + O(\stepsize_1)
\end{align*}
\end{lemma}

Similarly, we have the following result for $\vC$.
\begin{thm}
\label{thm:kfac_C_part}
The product $\vC \vC^\top$ has a first-order accuracy of the KFAC update of $\big(\vS_C  + \lambda \vI \big)^{-1}$  at each iteration if  the update of $\vC$ is updated according to Figure~\ref{fig:matDL_opt2} with the truncation of the matrix exponential and
these two updates use the same initialization and the same sequence of curvature matrices $\vG$.
\begin{align*}
    {\vC}   {\vC} ^\top      =   \big( \vS_C + \lambda \vI \big)^{-1}     + O(\stepsize_1^2)
\end{align*}
\end{thm}

\subsection{Proof of Lemma \ref{lemma:kfac_identity}}
We prove the lemma by induction
We first show the base case when $t=1$.
By definition, we have
\begin{align}
    \bar{\vS}_1 & = (1-\stepsize_1) \bar{\vS}_0 + \stepsize_1 \big( \hat{\vU}_0 +\lambda \vI \big) \\
    & = (1-\stepsize_1)  \hat{\vK}_0^{-T} \hat{\vK}_0^{-1} +  \stepsize_1 \big( \hat{\vU}_0 + \lambda \vI \big) \\
    &= \hat{\vK}_0^{-T} \Big[ \vI + \stepsize_1 \underbrace{ \Big(  \hat{\vK}_0^{T}  \hat{\vU}_0 \hat{\vK}_0 + \lambda \hat{\vK}_0^{T} \hat{\vK}_0  -\vI \Big)}_{=\hat{\vN}_0 }  \Big] \hat{\vK}_0^{-1} \\
    &= \hat{\vK}_0^{-T} \left[ \vI + \stepsize_1  \hat{\vN}_0  \right] \hat{\vK}_0^{-1}
\end{align}
Thus, the claim holds when $t=1$.

Suppose, the claim holds when $t=n$. By the claim, we have
\begin{align}
 \bar{\vS}_{n} =   \hat{\vK}_0^{-T} \left( \vI + \stepsize_1 \sum_{i=0}^{n-1}  \hat{\vN}_i  \right) \hat{\vK}_0^{-1} + O(\stepsize_1^2)
\end{align}
Now, we consider the case when $t=n+1$.
Notice that
\begin{align*}
    (1-\stepsize_1) \bar{\vS}_{n} & = \hat{\vK}_0^{-T} \left( \vI + \stepsize_1 \sum_{i=0}^{n-1}  \hat{\vN}_i  - \stepsize_1 \vI + O(\stepsize_1^2)  \right) \hat{\vK}_0^{-1} + O(\stepsize_1^2) \\
    &= \hat{\vK}_0^{-T} \left( \vI + \stepsize_1 \sum_{i=0}^{n-1}  \hat{\vN}_i - \stepsize_1 \vI   \right) \hat{\vK}_0^{-1} + O(\stepsize_1^2)
\end{align*}
By the definition of $\hat{\vS}_{n+1}$, we have
\begin{align}
   \bar{\vS}_{n+1} & = (1-\stepsize_1) \bar{\vS}_{n} + \stepsize_1 \big( \hat{\vU}_n + \lambda\vI \big) \\
   &= \hat{\vK}_0^{-T} \left( \vI + \stepsize_1 \sum_{i=0}^{n-1}  \hat{\vN}_i \underbrace{- \stepsize_1 \vI + \stepsize_1 \hat{\vK}_0^{T}
 \hat{\vU}_n  \hat{\vK}_0
+ \stepsize_1 \lambda \hat{\vK}_0^{T}
  \hat{\vK}_0
 }_{=\stepsize_1 \hat{\vN}_n} \right) \hat{\vK}_0^{-1}   + O(\stepsize_1^2) \\
 &= \hat{\vK}_0^{-T} \left( \vI + \stepsize_1 \sum_{i=0}^{n}  \hat{\vN}_i    \right) \hat{\vK}_0^{-1}   + O(\stepsize_1^2)
\end{align} which is exactly the claim when $t=n+1$.

Thus, by induction, the claim holds.

\subsection{Proof of Lemma \ref{lemma:ours_identity}}
We prove the lemma by induction
We first show the base case when $t=1$.
By definition, we have
\begin{align}
   \vK_1 = \vK_0 \Big( \vI - \frac{\stepsize_1}{2} \underbrace{\left( \vK_0^\top  \vU_0 \vK_0 + \lambda \vK_0^\top \vK_0   -\vI \right)}_{= \vN_0 } \Big)
\end{align}
Thus, the claim holds when $t=1$.

Suppose, the claim holds when $t=n$. By the claim, we have
\begin{align}
   \vK_n = \vK_0 \left( \vI - \frac{\stepsize_1}{2}
   \sum_{i=0}^{n-1} \vN_i  \right) + O(\stepsize_1^2)
\end{align}
Now, we consider the case when $t=n+1$.
Notice that
\begin{align}
   \vK_{n+1} & =   \vK_{n} \Big( \vI - \frac{\stepsize_1}{2}  \underbrace{ \Big( \vK_n^\top  \vU_n \vK_n + \lambda \vK_n^\top \vK_n -\vI \Big)}_{=\vN_n}  \Big) \\
   &= \underbrace{ \vK_0 \left( \vI - \frac{\stepsize_1}{2}
   \sum_{i=0}^{n-1} \vN_i  \right)}_{ = \vK_n -O(\stepsize_1^2) } \Big( \vI - \frac{\stepsize_1}{2} \vN_{n}  \Big)  + O(\stepsize_1^2) \\
   &=   \vK_0 \left( \vI - \frac{\stepsize_1}{2}
   \sum_{i=0}^{n-1} \vN_i  -  \frac{\stepsize_1}{2} \vN_{n} + O(\stepsize_1^2) \right)  + O(\stepsize_1^2) \\
    &=   \vK_0 \left( \vI - \frac{\stepsize_1}{2}
   \sum_{i=0}^{n} \vN_i  \right)  + O(\stepsize_1^2)
\end{align} which is exactly the claim when $t=n+1$.

Thus, by induction, the claim holds.

Notice that $\vN_i$ by definition is symmetric.
It is easy to see that
\begin{align}
    \vK_{t} \vK_{t}^\top &=  \vK_0 \left( \vI - \frac{\stepsize_1}{2}
   \sum_{i=0}^{t-1} \vN_i  \right) \left( \vI - \frac{\stepsize_1}{2}
   \sum_{i=0}^{t-1} \vN_i  \right)^\top  \vK_0^\top   + O(\stepsize_1^2) \\
   &= \vK_0 \left( \vI - \frac{\stepsize_1}{2}
   \sum_{i=0}^{t-1} \vN_i  \right) \left( \vI - \frac{\stepsize_1}{2}
   \sum_{i=0}^{t-1} \vN_i  \right)    \vK_0^\top   + O(\stepsize_1^2) \\
   &= \vK_0 \left( \vI -  \stepsize_1
   \sum_{i=0}^{t-1} \vN_i  \right) \vK_0^\top + O(\stepsize_1^2)
\end{align} Thus, the claim also holds.

\subsection{Proof of Lemma \ref{lemma:kfac_and_ours}}
We first show the base case when $t=0$.
By the assumption, we have $\vK_0 =  \hat{\vK}_0$.
Similarly, we have $\vU_0=\hat{\vU}_0$ by the assumption.

By definition, we have
\begin{align}
 \vN_0
  &=   {\vK}_0^\top  {\vU}_0  {\vK}_0 + \lambda  {\vK}_0^\top   {\vK}_0 - \vI  \\
 &= \hat{\vK}_0^\top \hat{\vU}_0 \hat{\vK}_0 + \lambda \hat{\vK}_0^\top   \hat{\vK}_0  - \vI \\
 & =\hat{\vN}_0
\end{align}
Thus, the claim holds when $t=0$.

When $t>0$, we can use Lemma \ref{lemma:ours_identity} to obtain the claim.
Notice that
\begin{align}
   {\vN}_{n+1}
    & = \vK_{n+1}^\top \vU_{n+1} \vK_{n+1} + \lambda \vK_{n+1}^\top \vK_{n+1} -\vI  \\
   &=  \left( \vI - \frac{\stepsize_1}{2}
   \sum_{i=0}^{n} \vN_i  \right)^\top  \vK_0^\top \big( \vU_{n+1} + \lambda \vI \big)  \vK_0 \left( \vI - \frac{\stepsize_1}{2}
   \sum_{i=0}^{n} \vN_i  \right) -\vI  + O(\stepsize_1^2)   \,\, \text{(Lemma 2)}\\
   &=     \vK_0^\top \big( \vU_{n+1} + \lambda \vI)  \vK_0  + O(\stepsize_1)  + O(\stepsize_1^2) \\
   &=  \hat{\vK}_0^\top  \big( \hat{\vU}_{n+1}  +\lambda \vI \big) \hat{\vK}_0 + O(\stepsize_1)\,\, \text{(Assumption)} \\
   &=  \hat{\vN}_{n+1}  + O(\stepsize_1)
\end{align}

\subsection{Proof of Theorem \ref{thm:kfac_K_part}}

It is sufficient to show that the following claim holds at iteration $t$ since $\bar{\vS}_t$ is non-singular.
\begin{align*}
    {\vK}_t   {\vK}^\top_t \bar{\vS}_t = \vI +  O(\stepsize_1^2)
\end{align*} where we use $\bar{\vS}_t$ to denote $\bar{\vS}_K$ at iteration $t$.

By assumptions, we know that Lemmas \ref{lemma:kfac_identity}, \ref{lemma:ours_identity},  \ref{lemma:kfac_and_ours} hold.
Moreover, we have $\vK_0=\hat{\vK}_0$.
Thus, we have
\begin{align}
     {\vK}_t   {\vK}^\top_t \bar{\vS}_t &= \vK_0 \left(\vI - \stepsize_1 \sum_{i=0}^{t-1} \vN_i \right) \vK_0^\top   \bar{\vS}_t + O(\stepsize_1^2) \text{ (by Lemma \ref{lemma:ours_identity}) }\\
     &= \vK_0 \left(\vI - \stepsize_1 \sum_{i=0}^{t-1} \vN_i \right) \vK_0^\top  \hat{\vK}_0^{-T} \left(\vI + \stepsize_1 \sum_{i=0}^{t-1} \hat{\vN}_i \right)  \hat{\vK}_0^{-1} + O(\stepsize_1^2) \text{ (by Lemma \ref{lemma:kfac_identity}) }\\
      &= \hat{\vK}_0 \left(\vI - \stepsize_1 \sum_{i=0}^{t-1} \hat{\vN}_i +O(\stepsize_1^2) \right)   \left(\vI + \stepsize_1 \sum_{i=0}^{t-1} \hat{\vN}_i \right)  \hat{\vK}_0^{-1} + O(\stepsize_1^2) \text{ (by Lemma \ref{lemma:kfac_and_ours}) } \\
      &=  \hat{\vK}_0 \vI  \hat{\vK}_0^{-1} + O(\stepsize_1^2) \\
      &= \vI + O(\stepsize_1^2)
\end{align}

\section{ Invariance
of INGD and SINGD
}
\label{app:invariance}

INGD and SINGD are scale invariant to the choice of the Kronecker approximation while KFAC and IKFAC are not.
Recall that we use the following Kronecker approximation to approximate the Hessian.
\begin{align*}
    \vU \otimes \vG \approx \nabla_\mu^2 \ell(\vmu)
\end{align*}
However, such an approximation is not unique. We can consider an equivalent approximation such as
\begin{align*}
    ( \alpha \vU ) \otimes  (\alpha^{-1} \vG ) \approx \nabla_\mu^2 \ell(\vmu)
\end{align*} where $\alpha \neq 0$ can be any arbitrary  non-zero scalar.

INGD is invariant  since the update scheme involving the  approximation is scale invariant: $\mathrm{Tr}(\vH_C) \vH_K = \mathrm{Tr}(\vC^T \vG \vC ) \vK^T \vU \vK = \mathrm{Tr}(\vC^T (\alpha^{-1} \vG) \vC ) \vK^T (\alpha \vU) \vK   $. The invariance is also preserved in SINGD since structures and their subspace projection maps are closed under scalar multiplications.

In contrast, the updates of KFAC and IKFAC are not scale invariant.
As an example, we consider using curvature approximations $\vU$ and $(\alpha \vU)$ to update $\vS_K^{-1}$ in KFAC, and denote the updated  $\vS_K^{-1}$
by $\hat{\vS}_K^{-1}$ and $\bar{\vS}_K^{-1}$, respectively.
As shown below,
we cannot recover
$\hat{\vS}_K^{-1}$ from $\bar{\vS}_K^{-1}$ by scale transformations and thus, the KFAC update is not scale invariant.
\begin{align*}
\hat{\vS}_K^{-1} = \big[ (1-\stepsize_1)
\hat{\vS}_K + \stepsize_1 \vU + \lambda \vI  \big]^{-1} \neq   \big[ (1-\stepsize_1)   \bar{\vS}_K + \stepsize_1 (\alpha \vU) + \lambda  \vI  \big]^{-1}  =   \bar{\vS}_K^{-1}
\end{align*}
An attempt to make the update of $\vS_K$ invariant is to set the damping weight to be $\alpha \lambda$.
However, the update of $\vS_C$ requires us to set the damping weight to be $\alpha^{-1} \lambda$ as shown below. Thus, it is impossible to make KFAC invariant without introducing individual damping weights.
\begin{align*}
\hat{\vS}_C^{-1} = \big[ (1-\stepsize_1)
\hat{\vS}_C + \stepsize_1 \vG + \lambda \vI  \big]^{-1} \neq   \big[ (1-\stepsize_1)   \bar{\vS}_C + \stepsize_1 (\alpha^{-1} \vG) + \lambda  \vI  \big]^{-1}  =   \bar{\vS}_C^{-1}
\end{align*}

\end{document}